%% file: main.tex
\documentclass[lettersize,journal]{IEEEtran}

\usepackage[utf8]{inputenc}
\usepackage[T1]{fontenc}
\usepackage{amsmath,amsfonts,amssymb}
\usepackage{nicefrac}
\usepackage{microtype}
\usepackage{graphicx}
\usepackage[table]{xcolor}
\usepackage{booktabs}
\usepackage{array}
\usepackage{multirow}
\usepackage{tabularx}
\newlength{\resultentrywidth}
\usepackage[caption=false,font=normalsize,labelfont=sf,textfont=sf]{subfig}
\usepackage{url}
\usepackage{cite}
\usepackage{pifont}
\usepackage{stfloats}
\newcounter{algorithm}
\renewcommand{\thealgorithm}{\arabic{algorithm}}
\usepackage{algorithmic}
\usepackage[colorlinks=true,linkcolor=blue,citecolor=blue,urlcolor=blue]{hyperref}

\newcommand{\cmark}{\textcolor[rgb]{0.0, 0.5, 0.0}{\ding{51}}}
\newcommand{\xmark}{\textcolor{red}{\ding{55}}}
\input{table_styles}
\input{method_names}
\input{appendix/layout}
\newcommand{\appendixref}[2]{%
  \ifcsname r@#1\endcsname\ref{#1}\else#2\fi}

\begin{document}

\title{ReLATE: Learning Relative Representations\\for Post-Hoc Multimodal Alignment}

\author{Shiwon~Kim, JaeSeong Hong and Yu~Rang~Park%
\thanks{S. Kim, J. Hong, and Y. R. Park are with the Department of Biomedical Systems Informatics, Yonsei University College of Medicine, Seoul 03722, Republic of Korea (e-mail: shiwon@yuhs.ac; jayhong7200@yuhs.ac; yurangpark@yuhs.ac).}%
\thanks{Corresponding author: Yu Rang Park.}%
\thanks{This work has been submitted to the IEEE for possible publication. Copyright may be transferred without notice, after which this version may no longer be accessible.}%
}

\markboth{Preprint}%
{Kim \MakeLowercase{\textit{et al.}}: ReLATE: Learning Relative Representations for Post-Hoc Multimodal Alignment}

\maketitle

\begin{abstract}
\input{sections/00_abstract}
\end{abstract}


\input{sections/01_introduction}
\input{sections/02_related_work}
\input{sections/03_preliminaries}
\input{sections/04_proposed_method}
\input{sections/05_experiments}
\input{sections/06_conclusion}

\bibliographystyle{IEEEtran}
\bibliography{ref}

\clearpage
\onecolumn
\markboth{Appendix}{Kim \MakeLowercase{\textit{et al.}}: Appendix}
\input{appendix/title_page}
\appendices
\setcounter{secnumdepth}{3}
\input{appendix/content}

\end{document}

%% file: table_styles.tex
\definecolor{relatehighlight}{HTML}{E8F3FA}

%% file: method_names.tex
\newcommand{\methodfull}{\textbf{Re}lative Alignment with \textbf{L}earnable \textbf{A}nchors and \textbf{T}oken \textbf{E}vidence}
\newcommand{\methodshort}{ReLATE}

%% file: appendix/layout.tex
\usepackage{changepage}
\newlength{\appendixtextmargin}
\newenvironment{appendixtext}{%
  \begin{adjustwidth}{\appendixtextmargin}{\appendixtextmargin}%
}{%
  \end{adjustwidth}%
}

%% file: sections/00_abstract.tex
Post-hoc multimodal alignment connects independently pretrained vision and language encoders while keeping them frozen, offering an alternative to joint pretraining when paired data are scarce. A central design challenge is to construct comparable representations across heterogeneous feature spaces while retaining local information relevant to distinct semantic concepts within each input. We propose \methodfull{} (\methodshort{}), which learns modality-specific anchors that jointly define relative representations and determine token contributions to each relative coordinate. Image patches and text tokens are represented by their similarities to anchors within their respective frozen feature spaces. Anchor-wise Similarity Pooling (ASP) then weights and aggregates these similarities separately for each anchor to construct the corresponding relative coordinate. A sample-level symmetric contrastive objective aligns the resulting image and text representations by optimizing only the anchors, without updating the encoders. Experiments on 13 general-domain benchmarks across four vision--language tasks demonstrate improvements over the compared post-hoc alignment methods under matched paired-data budgets, with further evaluations across data scales and encoder combinations. Beyond the general domain, we evaluate \methodshort{} in pathology as a specialized-domain application. By aligning frozen pathology image and biomedical language encoders, \methodshort{} matches or exceeds the evaluated pathology-specific vision--language models in balanced accuracy on three zero-shot classification benchmarks.

%% file: sections/01_introduction.tex
\section{Introduction}
\label{sec:intro}

\IEEEPARstart{V}{ision--language} models (VLMs) such as CLIP~\cite{radford2021learning} and ALIGN~\cite{jia2021scaling} have enabled a range of vision--language tasks, including zero-shot image classification and cross-modal retrieval, through joint learning from images and text. Developing these capabilities from scratch, however, requires large collections of image--text pairs and substantial computational resources. Obtaining sufficient paired data for multimodal pre-training is particularly challenging in specialized domains such as pathology~\cite{lu2024conch, xiang2025musk}. Meanwhile, powerful unimodal encoders are available for both visual and textual data~\cite{oquab2023dinov2, liu2019roberta}, including models tailored to specialized domains~\cite{chen2024uni, gu2021pubmedbert}.

The availability of these encoders has motivated methods that connect independently pretrained image and text models while keeping their parameters frozen~\cite{norelli2023asif, csa2025, vouitsis2024fusemix, freezealign, sail2025, structure2025}. These methods establish comparable multimodal representations by training compact alignment modules or using paired reference samples, enabling cross-modal comparison and zero-shot transfer without large-scale joint training. We refer to this process, performed after encoder pretraining, as \textit{post-hoc multimodal alignment}. Some of these methods use millions of image--text pairs~\cite{sail2025, freezealign}, while recent work achieves useful alignment with tens of thousands~\cite{vouitsis2024fusemix, structure2025}.

We consider two design questions for post-hoc multimodal alignment. First, such alignment requires a common coordinate system for representations produced by separately pretrained encoders. Previous approaches typically establish this system through learned projectors or adapters that map frozen features into a shared output space~\cite{vouitsis2024fusemix, freezealign, sail2025, structure2025}. Instead, we ask whether comparable coordinates can be defined through relationships computed within each frozen feature space. Second, existing methods often represent each input using a global feature or uniform aggregation of local features~\cite{csa2025, vouitsis2024fusemix, freezealign, sail2025, structure2025}. However, images and captions can contain multiple semantic concepts, and applying the same aggregation weights across coordinates may dilute concept-specific information. This raises the question of which tokens should contribute to each coordinate.

\begin{figure*}[!t]
    \captionsetup[subfloat]{font={normalsize,rm},labelfont=rm,textfont=rm}
    \centering
    \subfloat[Motivation behind \methodshort{}]{%
        \raisebox{6pt}{\includegraphics[width=0.55572\textwidth]{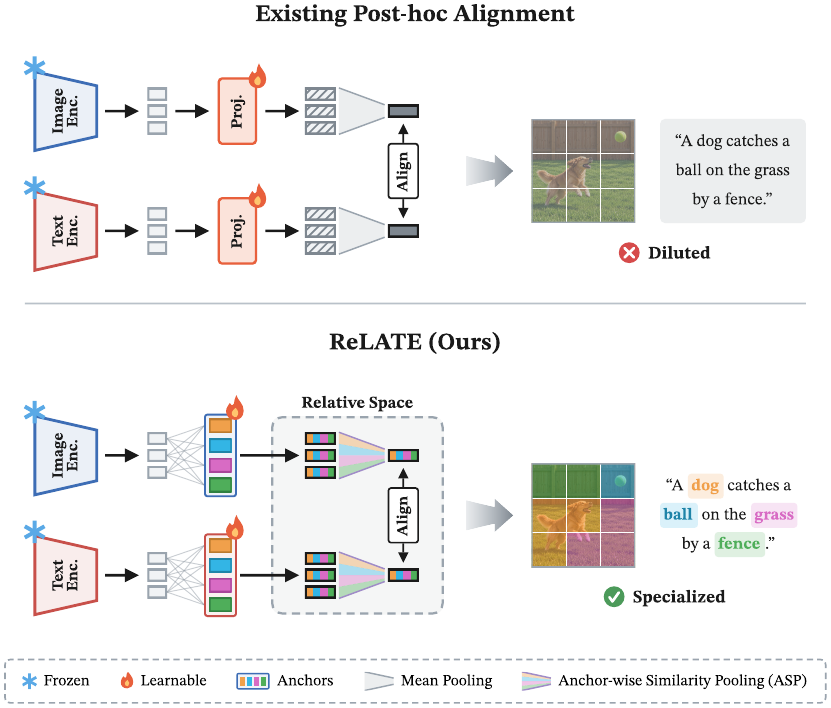}}%
        \label{fig:motivation_a}}
    \hfill
    \subfloat[Performance comparison]{%
        \raisebox{6pt}{\includegraphics[width=0.42428\textwidth]{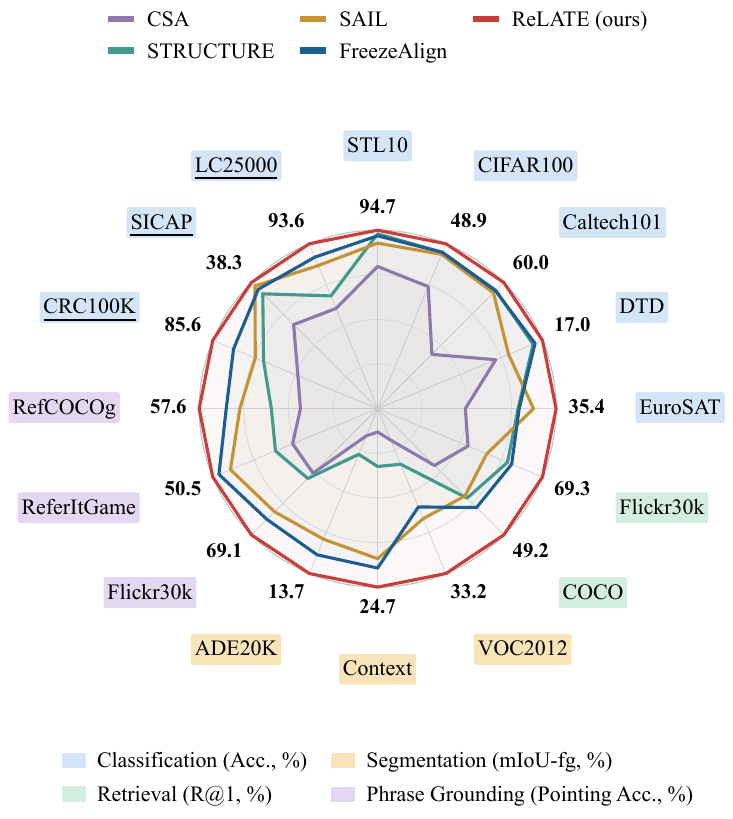}}%
        \label{fig:motivation_b}}
    \caption{(a) Top: mean pooling aggregates all tokens uniformly, so the evidence for each concept is diluted in the global representation. Bottom: \methodshort{} represents tokens in a shared relative space defined by modality-specific learnable anchors and aggregates token evidence separately for each anchor, so each coordinate becomes specialized to local cross-modal correspondence. (b) Performance across 16 benchmarks in general and pathology domains. Underlined dataset names indicate pathology benchmarks. Results are normalized on each axis for clarity, and the best raw scores (\%) are shown outside the radar. Retrieval scores average image-to-text and text-to-image.}
    \label{fig:motivation}
\end{figure*}

To address the first question, we turn to relative representations, which define coordinates through similarities to reference points within a feature space~\cite{moschella2023relrep}. Each reference point, termed an \textit{anchor}, defines one coordinate of the relative representation by measuring its similarity to an input embedding. Prior work connects independently trained spaces using embeddings of the same or paired reference samples as fixed anchors~\cite{moschella2023relrep, norelli2023asif, huang2024deepen}, motivated by the observation that relationships among inputs can be similar across models and modalities~\cite{huh2024platonic, moschella2023relrep}. However, anchors selected from sample embeddings are not necessarily well suited to the alignment objective. In our post-hoc setting, we therefore learn modality-specific anchors within the frozen feature spaces to optimize their positions and their correspondence for cross-modal alignment.

The second question is how to aggregate local information into these relative coordinates. Since each anchor defines one coordinate of the relative representation, we seek to construct each coordinate from tokens relevant to the corresponding anchor. We therefore compute token--anchor similarities before aggregation and use these values to weight token contributions separately for each anchor. Each coordinate thus summarizes the input from the perspective of its associated anchor. As illustrated in Fig.~\ref{fig:motivation}(a), existing post-hoc methods form global representations using global features or uniform aggregation of local features~\cite{freezealign, sail2025, structure2025}, which may dilute information relevant to individual concepts. In contrast, preserving anchor-specific token information in separate relative coordinates can encourage anchors to specialize in distinct semantic concepts and support local cross-modal correspondence.

We propose \methodfull{} (\textbf{\methodshort{}}), which combines modality-specific learnable anchors with \textit{Anchor-wise Similarity Pooling} (ASP) for post-hoc multimodal alignment. ASP assigns greater weight to tokens with higher similarity to each anchor and aggregates the token--anchor similarities into the corresponding relative coordinate. The resulting image and text representations are aligned through a sample-level symmetric contrastive objective that updates only the anchors in each modality space while keeping the encoders frozen. The anchors thus jointly determine the relative coordinates and the token contributions within them, without direct patch--word correspondence annotations.

We evaluate our method on 13 general-domain benchmarks spanning zero-shot classification, cross-modal retrieval, zero-shot segmentation, and phrase grounding. Evaluated under the same paired-data budgets, \methodshort{} outperforms the compared post-hoc alignment methods on all 13 benchmarks, with gains of up to 13.5 points in retrieval. We further study pathology as a practical specialized-domain application, connecting a frozen pathology image encoder with a biomedical language encoder. On three pathology benchmarks for zero-shot classification, \methodshort{} matches or exceeds the evaluated pathology-specific vision--language models. Fig.~\ref{fig:motivation}(b) summarizes performance across all 16 general-domain and pathology benchmarks. These results demonstrate the utility of turning existing unimodal encoders into vision--language models without jointly pretraining or fine-tuning the encoders.

Our contributions are threefold:
\begin{itemize}
    \item We propose \methodshort{}, a post-hoc multimodal alignment method that constructs comparable relative representations using modality-specific learnable anchors within frozen feature spaces. Rather than fixing the anchors to sample embeddings, we learn their positions and correspondence, enabling separately pretrained vision and language models to be connected without updating their parameters.
    \item To aggregate local information into these relative representations, we introduce Anchor-wise Similarity Pooling (ASP), which constructs each coordinate by weighting and aggregating token--anchor similarities separately for the corresponding anchor. The anchors thus determine both the relative coordinates and how tokens contribute to each coordinate. This design allows different coordinates to emphasize relevant local information and supports local cross-modal correspondence without direct patch--word correspondence annotations.
    \item We evaluate \methodshort{} on 13 general-domain benchmarks across four vision--language tasks and three pathology benchmarks for zero-shot classification. The results show improvements over the compared post-hoc methods under matched paired-data budgets and competitive performance against dedicated pathology vision--language models. We further examine performance across data scales and six encoder combinations, ablate the anchor and pooling configurations, and analyze cross-modal anchor consistency, specialization, and reuse.
\end{itemize}

%% file: sections/02_related_work.tex
\section{Related Work}
\label{sec:related}

\subsection{Post-hoc Multimodal Alignment}

Post-hoc multimodal alignment methods keep pretrained vision and language encoders frozen and connect their outputs for cross-modal comparison~\cite{norelli2023asif, csa2025, vouitsis2024fusemix, freezealign, sail2025, structure2025}. Training-free approaches establish cross-modal comparability using paired reference samples or canonical correlation analysis~\cite{norelli2023asif, csa2025}. Trainable methods fit alignment modules on frozen features. FreezeAlign~\cite{freezealign} and SAIL~\cite{sail2025} train projection modules for vision--language tasks, FuseMix~\cite{vouitsis2024fusemix} trains adapters on precomputed features with paired latent mixup, and STRUCTURE~\cite{structure2025} targets limited paired supervision with regularization that preserves multiscale neighborhood relations, together with similarity-based layer selection. These approaches also differ in how they aggregate encoder outputs into the final representation.

CSA~\cite{csa2025} and STRUCTURE~\cite{structure2025} operate on pooled global features, and FreezeAlign~\cite{freezealign} and SAIL~\cite{sail2025} combine the CLS feature with uniformly averaged patch features. These methods form global representations without making the token aggregation weights specific to individual output coordinates. \methodshort{} instead weights image patches and text tokens by their similarity to each learnable anchor and pools these similarities into the corresponding relative coordinate. This anchor-wise aggregation allows different coordinates to emphasize different semantic evidence within the same input.

\subsection{Relative Representations}

Relative representations describe an embedding through its similarities to a set of reference embeddings, or \textit{anchors}~\cite{moschella2023relrep}. The original work exploits the invariance of cosine similarity to angle-preserving transformations to enable zero-shot model stitching and latent space communication~\cite{moschella2023relrep}. Applications include cross-modal comparison using paired image and text references~\cite{norelli2023asif} and the construction of weakly aligned training data for vision--language pretraining~\cite{chen2023relit}. Relative spaces have also supported the combination of token distributions from heterogeneous LLMs in DEEPEN~\cite{huang2024deepen}, the reuse of encoder and policy components for new visual--task combinations in R3L~\cite{ricciardi2024r3l}, and semantic communication between models~\cite{huttebraucker2024semantic}.

Another line of work studies how to construct anchors and discover correspondences across spaces~\cite{huttebraucker2024semantic, cannistraci2023bootstrapping, svendsen2027param}. Semantic channel equalization constructs prototypical anchors by averaging corresponding samples within clusters~\cite{huttebraucker2024semantic}, while Bootstrapping Parallel Anchors estimates additional parallel anchors from a small set of known correspondences~\cite{cannistraci2023bootstrapping}. PARAM~\cite{svendsen2027param} learns anchors as convex mixtures of parallel support embeddings through mixture weights shared across spaces. Whereas these approaches construct anchors from support samples, \methodshort{} directly parameterizes anchors as learnable vectors within each frozen modality space and jointly optimizes them for vision--language alignment using paired supervision. These anchors also guide token aggregation: token--anchor similarities are computed before pooling, allowing the anchors to determine both the relative coordinates and token contributions to each coordinate.

%% file: sections/03_preliminaries.tex
\section{Preliminaries}
\label{sec:problem}

\subsection{Problem Formulation}

We consider independently pretrained image and text encoders whose weights remain frozen during alignment~\cite{freezealign, sail2025, structure2025}. Let $\mathcal D=\{(x_i^{v},x_i^{t})\}_{i=1}^{N}$ denote a paired image--text dataset, with modality index $m\in\{v,t\}$. For each input, the frozen encoder $f^{m}$ produces a sequence of token embeddings,
\begin{equation}
\mathbf Z_i^{m}=f^{m}(x_i^{m})
=\begin{bmatrix}
(\mathbf z_{i,1}^{m})^{\top}\\
\vdots\\
(\mathbf z_{i,T_i^{m}}^{m})^{\top}
\end{bmatrix}
\in\mathbb R^{T_i^{m}\times D_m},
\label{eq:frozen_tokens}
\end{equation}
where $T_i^{m}$ is the number of tokens and $D_m$ is the feature dimension. Each $\mathbf z_{i,u}^{m}\in\mathbb R^{D_m}$ corresponds to an image patch or a text token. The feature dimensions and sequence lengths need not be equal across modalities.

The goal is to construct modality-specific alignment functions $g^{v}$ and $g^{t}$ that map these encoder outputs to representations of a common dimension $d$,
\begin{equation}
\mathbf h_i^{m}=g^{m}(\mathbf Z_i^{m})\in\mathbb R^{d},
\qquad m\in\{v,t\},
\label{eq:alignment_mapping}
\end{equation}
so that images and texts can be compared directly using the same similarity function. Training encourages high similarity between paired image and text representations while retaining information that distinguishes different inputs. We use supervision only at the image and text sample level, without annotations specifying correspondences between individual image patches and text tokens.

\subsection{Relative Representations}

We formalize relative representations~\cite{moschella2023relrep} using an ordered set of anchors $\mathcal A^{m}=\{\mathbf a_k^{m}\}_{k=1}^{K}$, where $\mathbf a_k^{m}\in\mathbb R^{D_m}$. The relative representation of an embedding $\mathbf z\in\mathbb R^{D_m}$ is
\begin{equation}
\mathbf r^{m}(\mathbf z;\mathcal A^{m})
=\bigl[
\operatorname{cos}(\mathbf z,\mathbf a_1^{m}),
\ldots,
\operatorname{cos}(\mathbf z,\mathbf a_K^{m})
\bigr]^{\top}
\in\mathbb R^{K},
\label{eq:relative_coordinates}
\end{equation}
where
\begin{equation}
\operatorname{cos}(\mathbf z,\mathbf a)
=\frac{\mathbf z^{\top}\mathbf a}{\|\mathbf z\|_2\|\mathbf a\|_2}.
\label{eq:cosine_similarity}
\end{equation}

Each coordinate records the similarity to one anchor. Using the same number of anchors thus yields relative representations of equal dimension even when the original feature dimensions are not identical.

Comparing relative representations across spaces requires a correspondence between the anchors defining their coordinates~\cite{moschella2023relrep}. Such a correspondence can be established using fixed paired reference samples or learned through an alignment objective. \methodshort{} places learnable anchors in each modality space and jointly optimizes the two anchor sets to align the relative representations of paired image--text samples. The next section describes how token evidence is aggregated and aligned within the relative coordinates defined by these anchors.

%% file: sections/04_proposed_method.tex
\begin{figure*}[!t]
    \centering
    \begin{minipage}[t]{0.59\textwidth}
        \vspace{0pt}
        \centering
        \includegraphics[width=\linewidth]{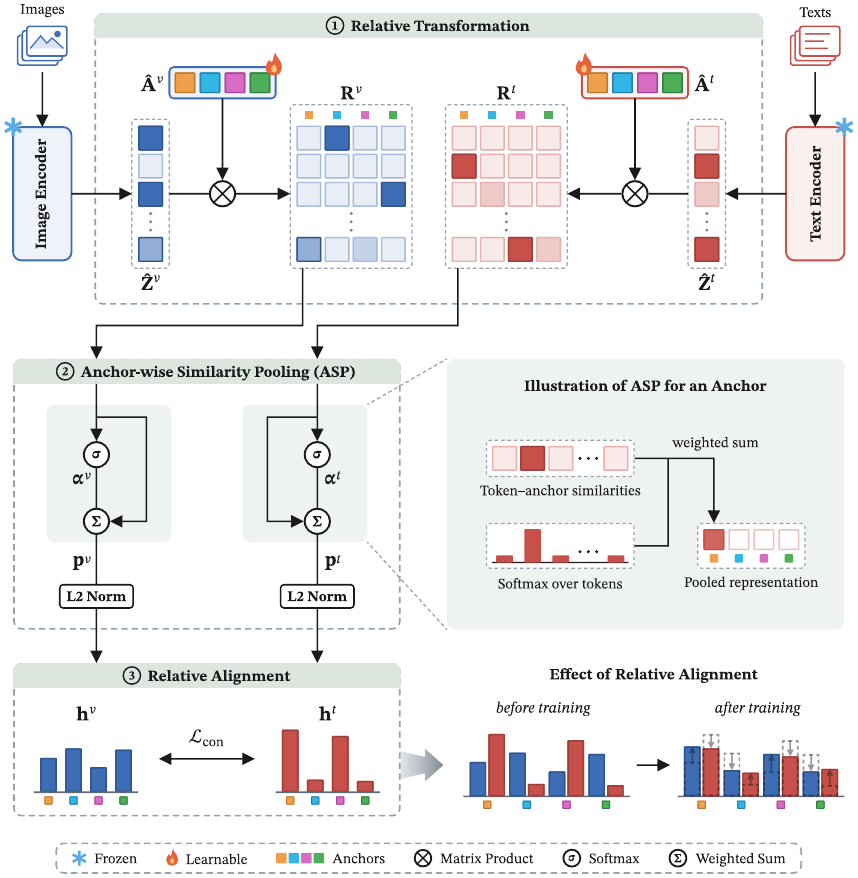}
        \caption{Overview of \methodshort{}. Frozen image and text tokens are transformed into relative representations using modality-specific anchors. Anchor-wise similarity pooling aggregates token-anchor similarities, and a symmetric contrastive loss aligns the pooled representations. For simplicity, the illustration uses four anchors per modality.}
        \label{fig:method}
    \end{minipage}\hfill
    \begin{minipage}[t]{0.39\textwidth}
        \vspace{0pt}
        \input{algorithms/01_training_procedure}
    \end{minipage}
\end{figure*}

\section{Proposed Method}
\label{sec:method}

\begingroup
\let\methoddisplayequation\equation
\renewcommand{\equation}{%
    \addtolength{\abovedisplayskip}{2pt}%
    \addtolength{\belowdisplayskip}{2pt}%
    \addtolength{\abovedisplayshortskip}{2pt}%
    \addtolength{\belowdisplayshortskip}{2pt}%
    \methoddisplayequation}

\methodshort{} learns cross-modal alignment by expressing frozen image and text features relative to modality-specific learnable anchors and aggregating token evidence separately for each anchor. As illustrated in Fig.~\ref{fig:method}, the computation comprises three stages: \textbf{(1) Relative Transformation} computes cosine similarities between each token and the modality-specific anchors; \textbf{(2) Anchor-wise Similarity Pooling} aggregates token similarities with anchor-dependent weights to form one relative representation per input; and \textbf{(3) Relative Alignment} aligns the representations of paired image--text samples using a symmetric contrastive objective. Both encoders remain frozen throughout training, and only the two modality-specific anchor matrices are optimized. Algorithm~\ref{alg:training} summarizes the training procedure.

\subsection{Relative Transformation}
\label{sec:relative_representation}

We apply the relative representation defined in Eq. \eqref{eq:relative_coordinates} to each token embedding of an input. For each modality $m\in\{v,t\}$, we introduce a matrix containing $K$ learnable anchor vectors as its rows:
\begin{equation}
\mathbf A^{m}=
\begin{bmatrix}
(\mathbf a_1^{m})^{\top}\\
\vdots\\
(\mathbf a_K^{m})^{\top}
\end{bmatrix}
\in\mathbb R^{K\times D_m}.
\label{eq:anchor_matrix}
\end{equation}

The two modalities use the same number of anchors, but each anchor lies in its respective encoder feature space, and the parameters are not shared across modalities. The anchors are initialized independently from a normal distribution and then individually $\ell_2$-normalized. The common index $k$ defines the relative coordinates compared across modalities.

For input $i$, let $\mathbf Z_i^{m}\in\mathbb R^{T_i^{m}\times D_m}$ denote the frozen token matrix, and let $\widehat{\mathbf Z}_i^{m}$ and $\widehat{\mathbf A}^{m}$ denote the row-wise $\ell_2$-normalized token and anchor matrices. We construct the token-level relative representation matrix by computing all token--anchor cosine similarities as follows:
\begin{equation}
\begin{aligned}
\mathbf R_i^{m}
&=\widehat{\mathbf Z}_i^{m}(\widehat{\mathbf A}^{m})^{\top}\\
&=\begin{bmatrix}
r_{i,1,1}^{m}&\cdots&r_{i,1,K}^{m}\\
\vdots&\ddots&\vdots\\
r_{i,T_i^{m},1}^{m}&\cdots&r_{i,T_i^{m},K}^{m}
\end{bmatrix}
\in\mathbb R^{T_i^{m}\times K},
\end{aligned}
\label{eq:rel_rep}
\end{equation}
where
\begin{equation}
r_{i,u,k}^{m}=[\mathbf R_i^{m}]_{u,k}
=\operatorname{cos}(\mathbf z_{i,u}^{m},\mathbf a_k^{m}).
\label{eq:token_anchor_similarity}
\end{equation}

Each row of $\mathbf R_i^{m}$ is the relative representation of one token, while each column records the responses of the token sequence to one anchor.

\subsection{Anchor-wise Similarity Pooling (ASP)}
\label{sec:cap}

To account for the multiple semantic elements within an input, we weight token contributions separately for each anchor. ASP applies softmax normalization to the token similarities for each anchor and uses the resulting weights to aggregate those similarities. Specifically, the weights for anchor $k$ are computed over the $T_i^{m}$ tokens as follows:
\begin{equation}
[\boldsymbol\alpha_i^{m}]_{u,k}
=\frac{\exp([\mathbf R_i^{m}]_{u,k}/\tau_p)}
{\displaystyle\sum_{u'=1}^{T_i^{m}}
\exp([\mathbf R_i^{m}]_{u',k}/\tau_p)},
\qquad u=1,\ldots,T_i^{m}.
\label{eq:cap_attention}
\end{equation}

Softmax is applied along the token dimension of each anchor column in $\mathbf R_i^{m}$, so the weights in each column sum to one. The pooling temperature $\tau_p>0$ controls their concentration: smaller $\tau_p$ concentrate weight on tokens with higher similarity to the anchor, whereas larger $\tau_p$ produce more uniform weights.

The $k$-th coordinate of the pooled relative representation is the weighted sum of similarities to the corresponding anchor:
\begin{equation}
[\mathbf p_i^{m}]_k
=\sum_{u=1}^{T_i^{m}}
[\boldsymbol\alpha_i^{m}]_{u,k}[\mathbf R_i^{m}]_{u,k},
\qquad \mathbf p_i^{m}\in\mathbb R^K.
\label{eq:cap_aggregation}
\end{equation}

Each coordinate summarizes token similarities to its anchor in a single value. Because the token weights are different across anchors, the final representation can emphasize different information within the input in different coordinates. These weights adapt as the anchors are learned. The final representation used for alignment is obtained by $\ell_2$-normalizing the pooled vector:
\begin{equation}
\mathbf h_i^{m}
=\frac{\mathbf p_i^{m}}{\|\mathbf p_i^{m}\|_2}
\in\mathbb R^K.
\label{eq:final_representation}
\end{equation}

\subsection{Relative Alignment}
\label{sec:objective}

For a mini-batch of $B$, we compute the cosine similarity of the normalized image representation $\mathbf h_i^v$ and text representation $\mathbf h_j^t$, scaled by a contrastive temperature $\tau>0$:
\begin{equation}
S_{ij}=\frac{(\mathbf h_i^{v})^{\top}\mathbf h_j^{t}}{\tau},
\qquad \mathbf S\in\mathbb R^{B\times B}.
\label{eq:alignment_logits}
\end{equation}

We use these scores to form a symmetric contrastive loss for cross-modal alignment in both directions~\cite{radford2021learning}:
\begin{equation}
\begin{split}
\mathcal L_{\mathrm{con}}
=-\frac{1}{2B}\sum_{i=1}^{B}
\Biggl[
&\underbrace{\log\frac{\exp(S_{ii})}{\sum_{j=1}^{B}\exp(S_{ij})}}_{\text{image-to-text}}\\
&+\underbrace{\log\frac{\exp(S_{ii})}{\sum_{j=1}^{B}\exp(S_{ji})}}_{\text{text-to-image}}
\Biggr].
\end{split}
\label{eq:contrastive_loss}
\end{equation}

This objective encourages the relative representations of paired image--text samples to have higher similarity than those of other samples in the batch. We jointly optimize the two anchor matrices by minimizing the following objective:
\begin{equation}
\min_{\mathbf A^{v},\mathbf A^{t}}\mathcal L_{\mathrm{con}}.
\label{eq:anchor_optimization}
\end{equation}

No patch--word correspondence annotations or local alignment losses are used.

\subsection{Inference}
\label{sec:inference}

After training, both the encoders and anchors are fixed. No further task-specific training is required.

\subsubsection{Classification and Retrieval}

Classification and retrieval use cosine similarity between the sample-level normalized representations $\mathbf h^v$ and $\mathbf h^t$. Given an image or text query, retrieval uses these scores to rank candidates from the other modality. Zero-shot classification compares an image with class text prototypes, formed by averaging and renormalizing each class's text prompt representations.

\subsubsection{Segmentation and Phrase Grounding}

For segmentation and phrase grounding, we use the relative representation of each spatial patch individually rather than pooling the image sequence into a single vector. For a single image, the normalized relative representation of patch $u$ and its score for text query $q$ are computed as follows:
\begin{equation}
\boldsymbol\rho_u^{v}
=\frac{[\mathbf R^{v}]_{u,:}^{\top}}
{\|[\mathbf R^{v}]_{u,:}\|_2},
\qquad
s_{u,q}=(\boldsymbol\rho_u^{v})^{\top}\mathbf h_q^{t}.
\label{eq:local_scores}
\end{equation}

Here, $\mathbf h_q^t$ denotes the class text prototype for segmentation, or the normalized query phrase representation obtained through ASP for phrase grounding. The patch scores are arranged on a spatial grid and upsampled to the image resolution. Segmentation assigns each pixel the class with the highest score. Phrase grounding selects the pixel with the highest phrase score as the pointing location.

\endgroup

%% file: algorithms/01_training_procedure.tex
\refstepcounter{algorithm}\label{alg:training}
\hrule
\vspace{3pt}
\noindent\textbf{Algorithm \thealgorithm:} Training procedure of \methodshort{}
\par\vspace{3pt}\hrule\vspace{3pt}
\small\raggedright
\setlength{\baselineskip}{12.5pt}
\begin{algorithmic}[1]
\setlength{\itemsep}{4pt}
\REQUIRE {\renewcommand{\arraystretch}{1.25}\begin{tabular}[t]{@{}l@{}}
Paired dataset $\mathcal D=\{(x_n^{v},x_n^{t})\}_{n=1}^{N}$;\\
Batch size $B$; Frozen encoders $f^{v},f^{t}$;\\
Anchors $\mathbf A^{v},\mathbf A^{t}$; Pooling temperature $\tau_p$;\\
Contrastive temperature $\tau$
\end{tabular}}
\vspace{6pt}
\ENSURE Learned anchors for cross-modal alignment
\FOR{each training iteration}
\STATE Sample $\{(x_i^{v},x_i^{t})\}_{i=1}^{B}$ from $\mathcal D$
\vspace{6pt}
\FOR{$m\in\{v,t\}$}
    \vspace{6pt}
    \STATE \ding{172} \textit{Relative transformation}
    \STATE $\mathbf Z_i^{m}\gets f^{m}(x_i^{m})$ for all $i$
    \STATE Row-normalize $\mathbf Z_i^{m}$ to obtain $\widehat{\mathbf Z}_i^{m}$
    \STATE Row-normalize $\mathbf A^{m}$ to obtain $\widehat{\mathbf A}^{m}$
    \STATE $\mathbf R_i^{m}\gets\widehat{\mathbf Z}_i^{m}\bigl(\widehat{\mathbf A}^{m}\bigr)^{\top}$
    \vspace{6pt}
    \STATE \ding{173} \textit{Anchor-wise similarity pooling}
    \STATE $\boldsymbol\alpha_i^{m}\gets\operatorname{softmax}_{\mathrm{tokens}}(\mathbf R_i^{m}/\tau_p)$
    \STATE $[\mathbf p_i^{m}]_k\gets\sum_u[\boldsymbol\alpha_i^{m}]_{u,k}[\mathbf R_i^{m}]_{u,k}$
    \STATE $\mathbf h_i^{m}\gets\mathbf p_i^{m}/\|\mathbf p_i^{m}\|_2$
\ENDFOR
\vspace{6pt}
\STATE \ding{174} \textit{Relative alignment}
\STATE $\mathcal L_{\mathrm{con}}\gets\operatorname{ContrastiveLoss}\bigl(\{(\mathbf h_i^{v},\mathbf h_i^{t})\}_{i=1}^{B},\tau\bigr)$
\STATE Update $\mathbf A^{v},\mathbf A^{t}$ by minimizing $\mathcal L_{\mathrm{con}}$
\ENDFOR
\STATE \textbf{return} learned anchors $\mathbf A^{v},\mathbf A^{t}$
\end{algorithmic}
\vspace{3pt}\hrule

%% file: sections/05_experiments.tex
\section{Experiments}
\label{sec:exp}

\begin{table*}[!t]
\input{tables/01_classification_retrieval}

\par\vspace{\floatsep}
\input{tables/02_segmentation_grounding}

\end{table*}

\subsection{Experimental Setup}
\label{sec:exp_setup}
\subsubsection*{Datasets and Evaluation}
For general-domain alignment, we use the MS COCO 2014 training split~\cite{lin2014coco}, with one caption per image resulting in 82,783 image--text pairs. We evaluate four downstream tasks. Zero-shot classification uses STL10~\cite{coates2011analysis}, CIFAR100~\cite{krizhevsky2009learning}, Caltech101~\cite{fei2004learning}, DTD~\cite{cimpoi2014describing}, and EuroSAT~\cite{helber2019eurosat}, reporting top-1 accuracy. Cross-modal retrieval reports image-to-text and text-to-image recall@1 on Flickr30k~\cite{young2014image} and the COCO Karpathy test split (in-domain held-out)~\cite{karpathy2015deep}. Zero-shot segmentation uses PASCAL VOC2012~\cite{everingham2010pascal}, PASCAL-Context~\cite{mottaghi2014role}, and ADE20K~\cite{zhou2017ade20k}, reporting foreground mean intersection-over-union (mIoU-fg). Phrase grounding uses the test splits of Flickr30k Entities~\cite{plummer2015entities}, ReferItGame~\cite{kazemzadeh2014referitgame}, and RefCOCOg (in-domain)~\cite{mao2016generation}. We report pointing accuracy~\cite{akbari2019multilevel}, the percentage of queries whose highest-scoring location lies within a ground-truth box. Unless otherwise specified, trainable methods report the mean and standard deviation over three seeds. Training-free methods use a single evaluation on the full alignment set; for smaller subsets, results are reported over three random sampling seeds. Detailed evaluation protocols and prompts are provided in Appendix~\appendixref{sec:supp_experimental_setup}{A}.

\subsubsection*{Baselines}
We compare against post-hoc alignment methods CSA~\cite{csa2025}, STRUCTURE~\cite{structure2025}, SAIL~\cite{sail2025}, and FreezeAlign~\cite{freezealign}. All methods share frozen encoders, layer pairs, alignment data, and evaluation splits, and use a 512-dimensional alignment space, following prior work~\cite{structure2025}; this corresponds to $K{=}512$ for \methodshort{}. Baseline-specific hyperparameters follow the recommended settings in their official implementations. For segmentation and grounding, each method maps spatial patch features for comparison with text representations. More details on baseline configurations including parameter counts can be found in Appendix~\appendixref{sec:supp_experimental_setup}{A}.

\subsubsection*{Implementation Details}
The default image and text encoders are DINOv2 ViT-L/14~\cite{oquab2023dinov2} and RoBERTa-Large~\cite{liu2019roberta}, with both frozen throughout alignment. Following STRUCTURE~\cite{structure2025}, we select the image--text layer pair with the highest mutual-kNN representational similarity on 5,000 randomly sampled training pairs, yielding image block 23 and text layer 24. Images use a resolution of $224\times224$.

\methodshort{} uses $K{=}512$ anchors per modality, pooling temperature $\tau_p{=}0.03$, and contrastive temperature $\tau{=}0.05$. Trainable methods use AdamW~\cite{loshchilov2019decoupled} optimizer with batch size 4,096, betas $(0.9,0.95)$, weight decay $10^{-4}$, and gradient clipping at 1.0. Learning rates are selected separately for each method and run using the same range-test procedure, followed by cosine annealing, and training uses early stopping on the validation loss computed on the COCO 2014 validation split. Alignment training uses a single NVIDIA Quadro RTX 8000 GPU (48\,GB). Additional details are provided in Appendix~\appendixref{sec:supp_experimental_setup}{A}. Alternative encoders and pathology-specific settings are described in their respective experiment sections.

\input{tables/03_anchor_pooling_ablation}

\subsection{Main Results}
\label{sec:results}

This section reports the main performance comparison with the four baselines using DINOv2 ViT-L and RoBERTa-Large encoders and the full 82K alignment pairs. Table~\ref{tab:main_results} covers zero-shot classification and cross-modal retrieval. Table~\ref{tab:dense_results} covers zero-shot segmentation and phrase grounding, with qualitative grounding examples in Fig.~\ref{fig:grounding_qualitative}.

\subsubsection{Zero-Shot Classification and Cross-Modal Retrieval}
Table~\ref{tab:main_results} reports zero-shot classification and cross-modal retrieval. \methodshort{} achieves the highest mean score on all nine benchmarks, and its standard deviation is smaller than its margin over the strongest baseline in every case. In classification, the improvements over the strongest baseline on each dataset range from 0.8 to 4.5 percentage points and are largest on EuroSAT (4.5) and Caltech101 (3.7). The smallest gain is on DTD, where all methods remain below 18\%, consistent with the limited texture vocabulary of COCO captions. This dependence on the vocabulary of the alignment pairs motivates the specialized-domain experiments in Section~\ref{sec:pathology}. Classification is also the task in which the margins are smallest overall. Each class is represented by a prompt that names a single concept, so a representation formed by uniform token aggregation already captures most of the relevant evidence.

Retrieval shows a different picture. \methodshort{} achieves the highest recall@1 in both directions on both datasets, improving over the strongest baseline, FreezeAlign, by 13.5 and 12.3 percentage points for image-to-text and text-to-image retrieval on Flickr30k and by 12.0 and 9.3 points on COCO. These margins are roughly four times those observed in classification. A caption typically describes several objects and attributes at once, and distinguishing the matched caption from similar ones requires the evidence for each of these concepts. This is the situation in which uniform pooling dilutes local evidence, whereas ASP preserves it in separate coordinates. The gains are of similar size on COCO, which is held-out in-domain data, and on Flickr30k, which is evaluated zero-shot, indicating that the improvement is not an artifact of in-domain training.

\subsubsection{Zero-Shot Segmentation and Phrase Grounding}
Table~\ref{tab:dense_results} examines whether representations learned from sample-level supervision transfer to patch-level prediction. For these tasks, image patches are scored individually against the text representation, as in Eq.~\eqref{eq:local_scores}, without pooling the image sequence. \methodshort{} achieves the highest mIoU-fg on all three segmentation datasets. On VOC2012, it reaches 33.2, exceeding the strongest baseline, SAIL (22.2), by 11.0 points; the gains on Context and ADE20K are 2.6 and 1.6 points over FreezeAlign. The strongest baseline also changes from SAIL on VOC2012 to FreezeAlign on the other two datasets, while \methodshort{} ranks first on all three. Because patches are scored without ASP at inference, these gains indicate that the learned anchors themselves define coordinates that are spatially consistent across modalities, a property examined further in Section~\ref{sec:anchor_analysis}.

For phrase grounding, \methodshort{} achieves the highest pointing accuracy on all three datasets, improving over FreezeAlign by 8.5 points on Flickr30k Entities, 8.8 points on RefCOCOg, and 1.9 points on ReferItGame. These results are obtained from image--caption pairs alone, without box supervision. Fig.~\ref{fig:grounding_qualitative} shows phrase-conditioned similarity maps for selected examples from the three datasets. \methodshort{} produces responses concentrated on the queried object, with the highest-scoring location inside the ground-truth box, whereas the responses of SAIL and FreezeAlign spread over co-occurring content and their predicted locations fall outside the box. This spreading is the qualitative counterpart of the diluted evidence illustrated in Fig.~\ref{fig:motivation}(a).

\begin{figure*}[!t]
    \centering
    \includegraphics[width=0.95\textwidth]{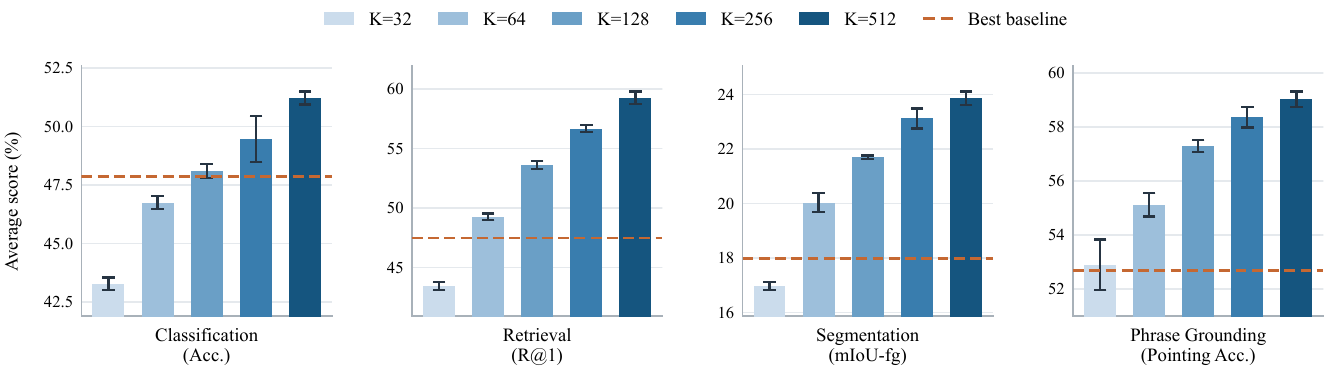}
    \caption{Effect of the number of anchors $K$ on task-averaged performance. Bars and error bars show the mean and standard deviation over three seeds. Dashed lines mark the strongest baseline task average: STRUCTURE for classification and FreezeAlign for the other tasks.}
    \label{fig:k_sweep}
\end{figure*}

\subsection{Ablation Study}
\label{sec:ablation}
We examine anchor formulations and pooling mechanisms, the effect of the number of anchors $K$, and sensitivity to the pooling temperature $\tau_p$.

\subsubsection{Anchor Formulation and Pooling}

Table~\ref{tab:pooling_ablation_summary} compares ASIF-based fixed anchors~\cite{norelli2023asif} with learnable anchors under mean, max, and ASP token pooling. Fixed anchors are evaluated at two budgets: $K{=}512$ paired samples, matching the number of learnable anchors, and the full alignment set ($K{=}82$K), matching the number of training pairs seen by the learnable anchors. Increasing the fixed-anchor budget improves all four tasks for every pooling mechanism. Nevertheless, \methodshort{}, which combines 512 learnable anchors with ASP, outperforms all fixed anchor variants. Appendix~\appendixref{app:anchor_learning_processing}{B} provides implementation details and additional fixed anchor comparisons.

The effect of pooling differs sharply between the two anchor settings. With fixed anchors, ASP does not outperform mean pooling at either budget. With learnable anchors, ASP improves over mean pooling on all four tasks, by 1.2, 10.9, 7.4, and 8.1 percentage points for classification, retrieval, segmentation, and grounding, respectively. This asymmetry follows from the design in Section~\ref{sec:cap}. The ASP weights are a function of token--anchor similarities, so they become selective only when the anchors respond to specific token evidence. A fixed anchor represents one reference sample as a whole rather than a concept that token selection could isolate, whereas a learnable anchor is optimized through ASP itself and adjusts to the token evidence it selects.

Neither component is sufficient on its own. Learnable anchors with mean pooling (50.0, 48.3, 16.5, and 50.9) perform at or below the level of learned-query multi-head attention (MHA) (49.7, 50.9, 18.7, and 54.4), which pools token features with a single learned query and a residual MLP head without anchors and thus represents learned non-uniform aggregation without a relative space. Also, ASP with fixed anchors does not improve on mean pooling. Only the combination of learnable anchors and ASP rises clearly above this level, demonstrating that the two components are complementary rather than independently beneficial. The MHA comparison further shows that the gain of ASP is not explained by attention-style weighting alone.

Max pooling yields the highest grounding average (59.5\%) but collapses in segmentation (7.2\% versus 23.9\% for ASP). It trains the anchors on the single most similar token per anchor, which suits pointing, where only the peak location matters, but leaves patch scores away from the peak poorly calibrated for dense labeling. ASP retains the grounding performance (59.0\%) while avoiding this failure.

\subsubsection{Number of Anchors $K$}

Fig.~\ref{fig:k_sweep} analyzes the number of anchors $K$, which sets both the dimensionality of the relative representation and the number of trainable parameters, $K(D_v+D_t)$. With DINOv2 ViT-L and RoBERTa-Large, this ranges from 0.07M at $K{=}32$ to 1.05M at $K{=}512$. Across the tested range of 32 to 512, task-averaged scores increase monotonically with $K$ on all four tasks. Retrieval and segmentation exceed the strongest baseline averages from $K{=}64$, where the anchors hold 0.13M parameters, roughly one tenth of the parameters of the smallest trainable baseline (Appendix~\appendixref{sec:supp_experimental_setup}{A}). Classification does so from $K{=}128$, and grounding shows a small advantage already at $K{=}32$ that widens with $K$.

\subsubsection{Pooling Temperature}
\input{tables/04_pooling_temperature}
Table~\ref{tab:tau_sweep_summary} examines the pooling temperature $\tau_p$, which controls the concentration of token weights for each anchor in ASP. Because the softmax weights approach uniform averaging as $\tau_p$ grows and a hard maximum as $\tau_p\to0$, this sweep interpolates between the mean and max pooling of Table~\ref{tab:pooling_ablation_summary}. Among the five tested values, $\tau_p=0.03$ achieves the highest mean score on all four tasks. Increasing $\tau_p$ from 0.03 to 0.10 reduces retrieval from 59.2\% to 52.9\% and segmentation from 23.9\% to 18.5\%, moving toward the mean-pooling results (48.3\% and 16.5\%). Grounding also declines, whereas classification varies less, consistent with the single-concept class prompts noted in Section~\ref{sec:results}. This trend reflects excessively smooth weights weakening the emphasis on anchor-relevant tokens. Conversely, the lower temperature of 0.02 does not improve over 0.03, and segmentation already drops to 21.8\%, in the direction of the max pooling collapse in Table~\ref{tab:pooling_ablation_summary}. We therefore use $\tau_p=0.03$ as the default setting.

\begin{figure*}[t]
    \centering
    \includegraphics[width=\textwidth]{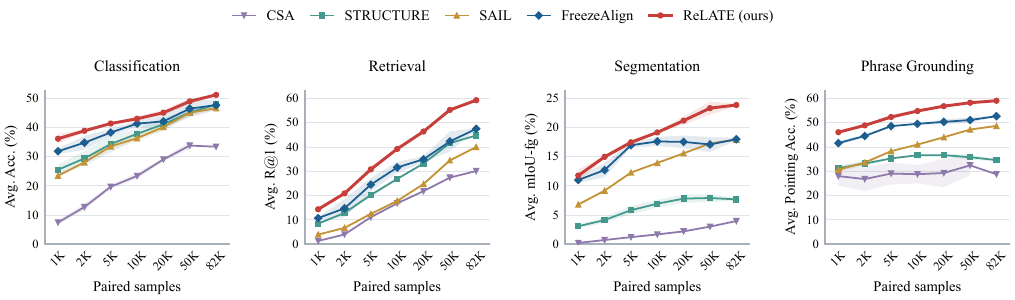}
    \caption{Performance as the number of paired alignment samples increases from 1K to 82K. Lines show mean task-averaged scores; shaded bands indicate standard deviation over three runs. CSA uses three sampled subsets at smaller sizes and a single full-data evaluation without an uncertainty band.}
    \label{fig:data_scaling}
\end{figure*}

\begin{figure*}[!t]
    \centering
    \includegraphics[width=\textwidth]{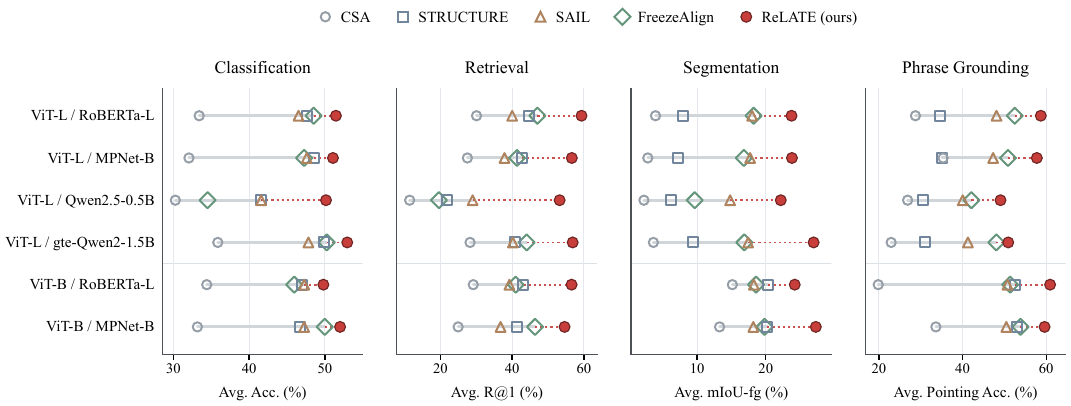}
    \caption{Task-averaged performance across six general-domain encoder pairs. Each row compares five alignment methods. Gray segments span the baseline scores; dotted red segments connect the strongest baseline to \methodshort{}, with the baseline selected separately for each task and encoder pair.}
    \label{fig:encoder_pairs}
\end{figure*}

\subsection{Scaling with Paired Data}
\label{sec:data_scaling}

Fig.~\ref{fig:data_scaling} examines how alignment performance depends on the amount of paired data, from 1K to the full alignment set of 82K pairs. \methodshort{} achieves the highest mean score on all four tasks at every evaluated data size. This holds already at 1K pairs, where the number of pairs is only twice the number of anchors, indicating that the learnable anchors do not overfit under scarce supervision. It also exceeds the strongest full-data baseline average using only 50K pairs for classification and retrieval and 10K pairs for segmentation and grounding. Moreover, the margin over the strongest baseline widens with the amount of paired data for retrieval, segmentation, and grounding, whereas it remains roughly constant for classification.

As the paired-data budget increases, the mean scores of \methodshort{} improve consistently on all four tasks, retrieval in particular, which still gains 4.0 points from 50K to 82K pairs. The baselines behave differently. On classification and retrieval, every method improves as the paired data grow, but the dense tasks separate the methods. FreezeAlign's segmentation score fluctuates within a narrow range beyond 5K pairs, and STRUCTURE's grounding score declines beyond 20K pairs. This divergence is consistent with what each method's supervision acts on. The baselines fit their modules to pooled global features, so patch-level scores are a by-product that additional pairs need not improve. In contrast, the pooled representation of \methodshort{} is built from the same token--anchor similarities used for patch-wise scoring, so additional pairs refine those similarities directly. Thus, the benefits of \methodshort{} extend beyond performance with scarce paired data to effective use of additional pairs within the evaluated range. Appendix~\appendixref{sec:supp_data_scaling}{C} further reports improvements in all four task averages when the alignment set is expanded to 110K pairs, and provides dataset-wise scaling curves.

\subsection{Generalization across Encoders}
\label{sec:encoder_generalization}

To examine whether the benefits of \methodshort{} depend on a particular encoder pair, we evaluate six combinations of frozen image and text models. DINOv2 ViT-L is paired with RoBERTa-Large, MPNet-Base~\cite{song2020mpnet}, Qwen2.5-0.5B~\cite{qwen2025qwen25}, and gte-Qwen2-1.5B~\cite{li2023gte}, while ViT-B is paired with RoBERTa-Large and MPNet-Base, as shown in Fig.~\ref{fig:encoder_pairs}. Alignment is trained separately for each pair using the same paired-data source. All combinations are evaluated with a single seed, including the default DINOv2 ViT-L and RoBERTa-Large pair, to keep the reporting consistent across pairs.

\methodshort{} achieves the highest task-averaged score for all six encoder pairs on each of the four tasks. The strongest baseline varies with both the encoder pair and the task, whereas the advantage of \methodshort{} is maintained across the evaluated combinations. The margin is largest with Qwen2.5-0.5B, a decoder-only causal language model not trained for sentence embeddings. Replacing RoBERTa-Large with Qwen2.5-0.5B lowers the retrieval average of FreezeAlign from 47.0\% to 19.6\% and that of STRUCTURE from 44.7\% to 21.8\%, whereas \methodshort{} drops from 59.3\% to 53.2\%. Classification shows the same pattern (FreezeAlign 48.5\% to 34.5\%, \methodshort{} 51.5\% to 50.1\%). \methodshort{} is thus considerably less sensitive to the choice of text model, suggesting that learning the reference points within each encoder space compensates for weaker text representations. Appendix~\appendixref{sec:supp_encoder_pairs}{D} provides complete dataset-wise results on all four tasks.

\begin{figure*}[!t]
\centering
\begin{minipage}[t]{0.49\textwidth}
\vspace{0pt}
\centering
\includegraphics[width=\linewidth]{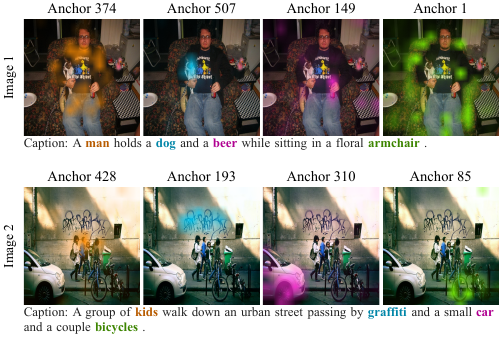}
\par\smallskip
{\normalsize (a) Anchor specialization}
\end{minipage}\hfill
\begin{minipage}[t]{0.49\textwidth}
\vspace{0pt}
\centering
\includegraphics[width=\linewidth]{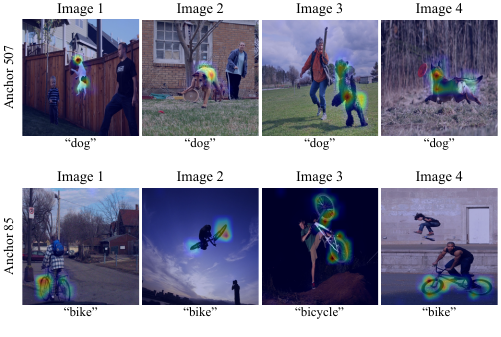}
\par\smallskip
{\normalsize (b) Anchor reuse}
\end{minipage}
\caption{Visualization of anchor specialization and reuse on Flickr30k. (a) Different anchors highlight image regions and caption words within two selected image--caption pairs; matching colors link the visual and textual responses of each anchor. (b) Responses of Anchor 507 to dogs and Anchor 85 to bicycles across four images each. The word below each image has the highest token attention weight for that anchor.}
\label{fig:qualitative}
\end{figure*}

\subsection{Application to Pathology Domain}
\label{sec:pathology}
\input{tables/05_pathology_classification}
We further evaluate pathology as a specialized-domain testbed for aligning independently pretrained unimodal encoders. All post-hoc methods align frozen UNI~\cite{chen2024uni} image and PubMedBERT~\cite{gu2021pubmedbert} text encoders using 220,318 image--text pairs from PathCap~\cite{sun2023pathasst}. Table~\ref{tab:pathology_results} reports zero-shot classification on CRC100K~\cite{kather2019crc}, SICAP~\cite{silva2020sicap}, and LC25000~\cite{borkowski2019lc25000} using balanced accuracy. All datasets average text representations over 22 histopathology templates~\cite{lu2024conch}; CRC100K additionally uses the class-name synonyms released with CONCH~\cite{lu2024conch}.

Under the same encoder and paired-data conditions, \methodshort{} achieves the highest mean balanced accuracy among the post-hoc methods on all three datasets, improving over the strongest baseline by 10.8, 1.1, and 7.6 percentage points on CRC100K, SICAP, and LC25000, respectively. The gain is largest on CRC100K, whose nine tissue-type classes are named frequently in pathology captions, and smallest on SICAP, where Gleason grading requires fine morphological distinctions that captions describe less consistently; all methods remain below 40\% on SICAP, mirroring the DTD result in Section~\ref{sec:results}. To assess practical utility, Table~\ref{tab:pathology_results} also includes the general-domain VLM CLIP~\cite{radford2021learning} and pathology-specific VLMs QuiltNet~\cite{ikezogwo2023quilt}, PLIP~\cite{huang2023plip}, CONCH~\cite{lu2024conch}, and MUSK~\cite{xiang2025musk} as references with different pretraining data and architectures. \methodshort{} attains the highest mean balanced accuracy among all compared models on each dataset, reaching 85.6\% on CRC100K compared with 79.9\% for the strongest pathology VLM, MUSK, and 27.1\% for general-domain CLIP. On SICAP and LC25000, \methodshort{} and the strongest pathology VLMs perform comparably, with margins of 1.1 and 0.7 points within one standard deviation. These results demonstrate the utility of reusing specialized unimodal encoders for vision--language tasks without updating their weights. Appendix~\appendixref{sec:supp_pathology_prompts}{E} provides additional metrics and benchmark performance, together with scaling experiments on paired pathology data.

\subsection{Analysis of Learned Anchors}
\label{sec:anchor_analysis}

\begin{figure}[!t]
    \centering
    \includegraphics[width=\columnwidth]{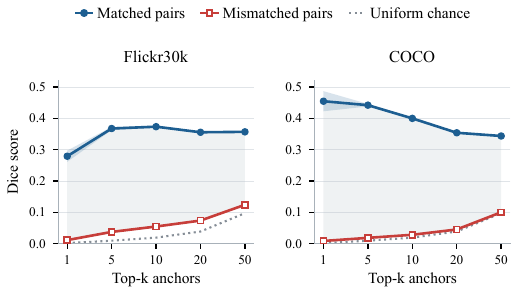}
    \caption{Cross-modal anchor consistency on Flickr30k and COCO test split, measured by Dice scores between the top-$k$ image and text anchor sets over three seeds. Dotted gray lines show the expected overlap under independent uniform anchor selection ($k/512$).}
    \label{fig:anchor_consistency}
\end{figure}

We analyze the behavior of learned anchors through visualizations in Fig.~\ref{fig:qualitative} and a quantitative evaluation of cross-modal consistency in Fig.~\ref{fig:anchor_consistency}. Together, these analyses examine the semantic roles of individual anchors and the correspondence between their image and text responses.

We first examine whether matched images and captions emphasize similar coordinates in the learned relative space. Fig.~\ref{fig:anchor_consistency} evaluates this cross-modal consistency on Flickr30k and COCO. For each sample $i$, we independently select the top-$k$ coordinates of the image and text representations $\mathbf h_i^v$ and $\mathbf h_i^t$ from $K{=}512$ anchors, and the Dice score~\cite{dice1945measures} measures the fraction of selected anchors shared between the two modalities. Matched pairs share far more anchors than randomly mismatched pairs on both datasets for every evaluated $k\in\{1,5,10,20,50\}$, and the mismatched scores stay close to the chance level $k/512$, so the overlap of matched pairs is not an artifact of a few anchors responding to every input. The effect is sharpest at small $k$: at $k=1$, the single most responsive anchor coincides for 45.5\% of matched pairs on COCO and 27.9\% on Flickr30k, against 0.9\% and 1.2\% for mismatched pairs; at $k=5$, the matched--mismatched gaps are 0.424 and 0.330. The gap narrows as $k$ grows because the chance level rises with $k$. Appendix~\appendixref{app:anchor_overlap}{F} provides the detailed evaluation protocol and additional overlap metrics, including Soft Dice and Jaccard scores.

To complement this quantitative analysis, Fig.~\ref{fig:qualitative}(a) examines which image regions and caption words individual anchors emphasize. Within each selected image--caption pair, different anchors exhibit distinct responses. In Image 1, the highlighted regions and words correspond to the man, dog, beer, and armchair; in Image 2, they correspond to the kids, graffiti, car, and bicycles. The correspondence between these visual responses and the highest-weighted caption words illustrates how separate anchor coordinates can emphasize different components of the same input.

Fig.~\ref{fig:qualitative}(b) examines anchor reuse across different image--caption pairs. Anchor 507 responds to dog-related regions and the word ``dog'' across four scenes, while Anchor 85 responds to bicycle-related regions with the caption words ``bike'' or ``bicycle.'' These examples illustrate that an anchor can retain a similar semantic role across changes in scene and appearance. Together, the two panels show complementary behaviors: different anchors emphasize distinct components within a sample, while the same anchor can be reused for related content across samples. Such selective responses emerge from image--caption alignment without assigning explicit concept labels to individual anchors.

%% file: tables/01_classification_retrieval.tex
\caption{Zero-shot classification (top-1 accuracy, \%) and cross-modal retrieval (recall@1, \%) results. Trainable methods report mean $\pm$ standard deviation over three seeds. Best and second-best results are shown in \textbf{bold} and \underline{underlined}, respectively.}
\label{tab:main_results}
\centering
\renewcommand{\arraystretch}{1.20}
\footnotesize
\setlength{\tabcolsep}{3pt}
\settowidth{\resultentrywidth}{\textbf{99.9}\,$\pm$\,9.9}
\begin{tabularx}{\textwidth}{l *{9}{>{\raggedright\arraybackslash\hspace*{\dimexpr(\hsize-\resultentrywidth)/2\relax}}X}}
\toprule
& \multicolumn{5}{c}{\textbf{Zero-shot Classification} (Acc., \%)}
& \multicolumn{4}{c}{\textbf{Cross-modal Retrieval} (R@1, \%)} \\
\cmidrule(lr){2-6} \cmidrule(lr){7-10}
\textbf{Method}
& \multicolumn{1}{c}{STL10} & \multicolumn{1}{c}{CIFAR100} & \multicolumn{1}{c}{Caltech101} & \multicolumn{1}{c}{DTD} & \multicolumn{1}{c}{EuroSAT}
& \multicolumn{1}{c}{Flickr I2T} & \multicolumn{1}{c}{Flickr T2I}
& \multicolumn{1}{c}{COCO I2T} & \multicolumn{1}{c}{COCO T2I} \\
\midrule
CSA~\cite{csa2025} & 75.5 & 36.2 & 25.8 & 12.2 & 17.4 & 43.4 & 32.6 & 25.1 & 19.2 \\
STRUCTURE~\cite{structure2025} & \underline{92.8}\,$\pm$\,0.1 & 46.3\,$\pm$\,0.3 & \underline{56.3}\,$\pm$\,0.6 & 16.0\,$\pm$\,0.2 & 27.9\,$\pm$\,2.0 & 61.3\,$\pm$\,0.7 & 47.8\,$\pm$\,0.3 & 40.4\,$\pm$\,0.1 & 29.4\,$\pm$\,0.2 \\
SAIL~\cite{sail2025} & 87.8\,$\pm$\,2.7 & 45.8\,$\pm$\,0.2 & 55.1\,$\pm$\,2.0 & 13.5\,$\pm$\,1.1 & \underline{30.9}\,$\pm$\,3.4 & 51.8\,$\pm$\,0.3 & 39.9\,$\pm$\,1.2 & 38.7\,$\pm$\,0.4 & 29.4\,$\pm$\,0.3 \\
FreezeAlign~\cite{freezealign} & 91.7\,$\pm$\,2.4 & \underline{46.4}\,$\pm$\,1.0 & 56.1\,$\pm$\,0.3 & \underline{16.2}\,$\pm$\,0.6 & 28.1\,$\pm$\,4.3 & \underline{63.3}\,$\pm$\,0.4 & \underline{49.4}\,$\pm$\,0.4 & \underline{43.8}\,$\pm$\,0.8 & \underline{33.3}\,$\pm$\,0.6 \\
\midrule
\rowcolor{relatehighlight}
\textbf{\methodshort{} (ours)} & \textbf{94.7}\,$\pm$\,0.8 & \textbf{48.9}\,$\pm$\,0.4 & \textbf{60.0}\,$\pm$\,0.9 & \textbf{17.0}\,$\pm$\,0.7 & \textbf{35.4}\,$\pm$\,0.9 & \textbf{76.8}\,$\pm$\,1.3 & \textbf{61.7}\,$\pm$\,0.3 & \textbf{55.8}\,$\pm$\,0.6 & \textbf{42.6}\,$\pm$\,0.2 \\
\bottomrule
\end{tabularx}

%% file: tables/02_segmentation_grounding.tex
\begin{minipage}[t]{\columnwidth}
\vspace{0pt}
\caption{Zero-shot segmentation (mIoU-fg, \%) and phrase grounding (pointing accuracy, \%) results. Trainable methods report mean $\pm$ standard deviation over three seeds. Best and second-best results are shown in \textbf{bold} and \underline{underlined}, respectively.}
\label{tab:dense_results}
\label{tab:segmentation}
\label{tab:grounding}
\centering
\renewcommand{\arraystretch}{1.20}
\footnotesize
\setlength{\tabcolsep}{3pt}
\vspace{1.8pt}
\settowidth{\resultentrywidth}{\textbf{99.9}\,$\pm$\,9.9}
\begin{tabularx}{\columnwidth}{l *{3}{>{\raggedright\arraybackslash\hspace*{\dimexpr(\hsize-\resultentrywidth)/2\relax}}X}}
\toprule
\multicolumn{4}{c}{\textbf{Zero-shot Segmentation} (mIoU-fg, \%)} \\
\midrule
\textbf{Method} & \multicolumn{1}{c}{VOC2012} & \multicolumn{1}{c}{Context} & \multicolumn{1}{c}{ADE20K} \\
\midrule
CSA~\cite{csa2025} & 6.3 & 3.2 & 2.3 \\
STRUCTURE~\cite{structure2025} & 11.2\,$\pm$\,0.5 & 8.0\,$\pm$\,0.2 & 3.8\,$\pm$\,0.1 \\
SAIL~\cite{sail2025} & \underline{22.2}\,$\pm$\,0.1 & 20.8\,$\pm$\,0.4 & 10.8\,$\pm$\,0.3 \\
FreezeAlign~\cite{freezealign} & 19.8\,$\pm$\,0.3 & \underline{22.1}\,$\pm$\,0.5 & \underline{12.1}\,$\pm$\,0.3 \\
\midrule
\rowcolor{relatehighlight}
\textbf{\methodshort{} (ours)} & \textbf{33.2}\,$\pm$\,1.4 & \textbf{24.7}\,$\pm$\,0.7 & \textbf{13.7}\,$\pm$\,0.1 \\
\midrule
\addlinespace[3pt]
\specialrule{\lightrulewidth}{0pt}{3pt}
\multicolumn{4}{c}{\textbf{Phrase Grounding} (Pointing Acc., \%)} \\
\midrule
\textbf{Method} & \multicolumn{1}{c}{Flickr30k} & \multicolumn{1}{c}{ReferItGame} & \multicolumn{1}{c}{RefCOCOg} \\
\midrule
CSA~\cite{csa2025} & 35.3 & 26.0 & 24.9 \\
STRUCTURE~\cite{structure2025} & 38.2\,$\pm$\,0.7 & 31.3\,$\pm$\,0.5 & 34.3\,$\pm$\,0.3 \\
SAIL~\cite{sail2025} & 56.4\,$\pm$\,0.6 & 45.1\,$\pm$\,1.1 & 44.4\,$\pm$\,0.2 \\
FreezeAlign~\cite{freezealign} & \underline{60.6}\,$\pm$\,0.3 & \underline{48.6}\,$\pm$\,0.4 & \underline{48.8}\,$\pm$\,0.2 \\
\midrule
\rowcolor{relatehighlight}
\textbf{\methodshort{} (ours)} & \textbf{69.1}\,$\pm$\,0.2 & \textbf{50.5}\,$\pm$\,0.4 & \textbf{57.6}\,$\pm$\,0.8 \\
\bottomrule
\end{tabularx}
\end{minipage}\hfill
\begin{minipage}[t]{\columnwidth}
\vspace{6pt}
\makeatletter\def\@captype{figure}\makeatother
    \centering
    \includegraphics[width=\columnwidth]{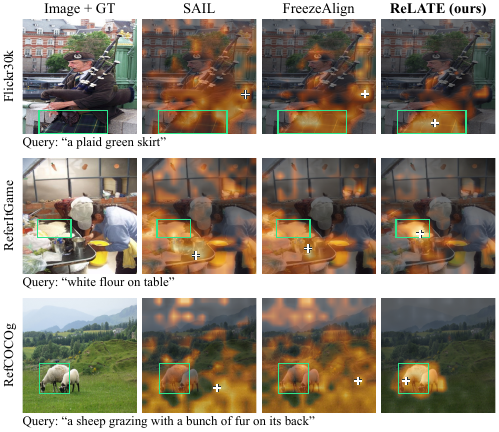}
    \par\vspace{-0.5pt}
    \caption{Phrase grounding visualizations. Green boxes indicate ground-truth regions, and white crosses mark predicted pointing locations. Heatmaps show cosine similarities between image patches and the query phrase.}
    \label{fig:grounding_qualitative}
\end{minipage}

%% file: tables/03_anchor_pooling_ablation.tex
\begin{table*}[!t]
\centering
\begin{minipage}{0.8\textwidth}
\caption{Anchor formulation and pooling ablations. Scores are task averages (\%). Fixed anchors are ASIF-based~\cite{norelli2023asif}. Learned-query multi-head attention (MHA) and learnable anchor results are mean $\pm$ standard deviation over three seeds; fixed anchor at $K{=}512$ uses three sampled subsets of anchors, while $K{=}82$K uses a single evaluation. Best results are in \textbf{bold}; second-best results are \underline{underlined}.}
\label{tab:pooling_ablation_summary}
\end{minipage}\par
\centering
\footnotesize
\renewcommand{\arraystretch}{1.20}
\setlength{\tabcolsep}{4pt}
\settowidth{\resultentrywidth}{\textbf{99.9}\,$\pm$\,9.9}
\begin{tabularx}{0.72\textwidth}{ll *{4}{>{\raggedright\arraybackslash\hspace*{\dimexpr(\hsize-\resultentrywidth)/2\relax}}X}}
\toprule
\textbf{Anchors} & \textbf{Pooling} & \multicolumn{1}{c}{\textbf{Avg. Cls.}} & \multicolumn{1}{c}{\textbf{Avg. Ret.}} & \multicolumn{1}{c}{\textbf{Avg. Seg.}} & \multicolumn{1}{c}{\textbf{Avg. Gr.}} \\
\midrule
-- & Learned-query MHA & 49.7\,$\pm$\,0.2 & 50.9\,$\pm$\,0.4 & \underline{18.7}\,$\pm$\,0.2 & 54.4\,$\pm$\,0.1 \\
\midrule
Fixed~\cite{norelli2023asif} & Mean & 29.0\,$\pm$\,1.8 & 3.9\,$\pm$\,0.3 & 11.9\,$\pm$\,0.6 & 39.4\,$\pm$\,0.5 \\
($K{=}512$) & Max & 23.6\,$\pm$\,1.5 & 3.1\,$\pm$\,0.3 & 11.6\,$\pm$\,0.8 & 39.2\,$\pm$\,0.5 \\
 & ASP & 23.9\,$\pm$\,1.8 & 2.9\,$\pm$\,0.4 & 11.5\,$\pm$\,0.9 & 39.3\,$\pm$\,0.5 \\
\midrule
Fixed~\cite{norelli2023asif} & Mean & 43.9 & 17.1 & 17.7 & 50.6 \\
($K{=}82$K) & Max & 36.2 & 11.0 & 17.1 & 50.1 \\
 & ASP & 36.1 & 10.6 & 17.1 & 50.0 \\
\midrule
Learnable & Mean & \underline{50.0}\,$\pm$\,0.9 & 48.3\,$\pm$\,0.0 & 16.5\,$\pm$\,0.1 & 50.9\,$\pm$\,0.4 \\
($K{=}512$) & Max & 48.4\,$\pm$\,0.5 & \underline{53.9}\,$\pm$\,0.4 & 7.2\,$\pm$\,0.6 & \textbf{59.5}\,$\pm$\,0.3 \\
 & \cellcolor{relatehighlight}\textbf{ASP (\methodshort{})} & \cellcolor{relatehighlight}\textbf{51.2}\,$\pm$\,0.3 & \cellcolor{relatehighlight}\textbf{59.2}\,$\pm$\,0.5 & \cellcolor{relatehighlight}\textbf{23.9}\,$\pm$\,0.3 & \cellcolor{relatehighlight}\underline{59.0}\,$\pm$\,0.3 \\
\bottomrule
\end{tabularx}
\end{table*}

%% file: tables/04_pooling_temperature.tex
\begin{table}[!t]
\caption{Sensitivity to the pooling temperature $\tau_p$. Scores are task averages (\%), mean $\pm$ standard deviation over three seeds. Best results are in \textbf{bold}; second-best results are \underline{underlined}.}
\label{tab:tau_sweep_summary}
\centering
\renewcommand{\arraystretch}{1.20}
\footnotesize
\setlength{\tabcolsep}{2pt}
\begin{tabularx}{\columnwidth}{>{\centering\arraybackslash}p{26pt} *{4}{>{\centering\arraybackslash}X}}
\toprule
\textbf{$\tau_p$} & \textbf{Avg. Cls.} & \textbf{Avg. Ret.} & \textbf{Avg. Seg.} & \textbf{Avg. Gr.} \\
\midrule
0.02 & \underline{51.0}\,$\pm$\,0.2 & \underline{58.8}\,$\pm$\,0.3 & 21.8\,$\pm$\,0.2 & \underline{58.6}\,$\pm$\,0.8 \\
\rowcolor{relatehighlight}
0.03 & \textbf{51.2}\,$\pm$\,0.3 & \textbf{59.2}\,$\pm$\,0.5 & \textbf{23.9}\,$\pm$\,0.3 & \textbf{59.0}\,$\pm$\,0.3 \\
0.05 & 50.5\,$\pm$\,0.2 & 57.6\,$\pm$\,0.4 & \underline{22.9}\,$\pm$\,0.3 & 57.5\,$\pm$\,0.0 \\
0.07 & 50.4\,$\pm$\,0.7 & 55.6\,$\pm$\,0.2 & 20.8\,$\pm$\,0.3 & 55.3\,$\pm$\,0.1 \\
0.10 & 50.1\,$\pm$\,1.2 & 52.9\,$\pm$\,0.1 & 18.5\,$\pm$\,0.1 & 53.3\,$\pm$\,0.3 \\
\bottomrule
\end{tabularx}
\end{table}

%% file: tables/05_pathology_classification.tex
\begin{table}[!t]
\caption{Zero-shot pathology classification (balanced accuracy, \%). Trainable post-hoc methods report mean $\pm$ standard deviation over three seeds; CSA and VLMs use single evaluations. Best results are in \textbf{bold}; second-best results are \underline{underlined}.}
\label{tab:pathology_results}
\centering
\footnotesize
\renewcommand{\arraystretch}{1.20}
\setlength{\tabcolsep}{2pt}
\settowidth{\resultentrywidth}{\textbf{99.9}\,$\pm$\,9.9}
\begin{tabularx}{\columnwidth}{l *{3}{>{\raggedright\arraybackslash\hspace*{\dimexpr(\hsize-\resultentrywidth)/2\relax}}X}}
\toprule
& \multicolumn{3}{c}{\textbf{Zero-shot Classification} (Balanced Acc.)} \\
\cmidrule(lr){2-4}
\textbf{Method} & \multicolumn{1}{c}{CRC100K$^{\dagger}$} & \multicolumn{1}{c}{SICAP} & \multicolumn{1}{c}{LC25000} \\
\midrule
CLIP~\cite{radford2021learning} & 27.1 & 20.2 & 27.7 \\
\midrule
QuiltNet~\cite{ikezogwo2023quilt} & 59.8 & 30.9 & 73.2 \\
PLIP~\cite{huang2023plip} & 67.4 & 27.1 & 77.4 \\
CONCH~\cite{lu2024conch} & 79.1 & 35.0 & \underline{92.9} \\
MUSK~\cite{xiang2025musk} & \underline{79.9} & \underline{37.2} & 88.8 \\
\midrule
CSA~\cite{csa2025} & 41.4 & 25.5 & 56.9 \\
STRUCTURE~\cite{structure2025} & 59.2\,$\pm$\,1.1 & 34.8\,$\pm$\,1.1 & 64.0\,$\pm$\,1.3 \\
SAIL~\cite{sail2025} & 63.5\,$\pm$\,2.6 & \underline{37.2}\,$\pm$\,4.0 & 81.2\,$\pm$\,3.4 \\
FreezeAlign~\cite{freezealign} & 74.8\,$\pm$\,6.2 & 36.2\,$\pm$\,3.2 & 86.0\,$\pm$\,2.0 \\
\midrule
\cellcolor{relatehighlight}\textbf{\methodshort{} (ours)} & \cellcolor{relatehighlight}\textbf{85.6}\,$\pm$\,1.9 & \cellcolor{relatehighlight}\textbf{38.3}\,$\pm$\,3.3 & \cellcolor{relatehighlight}\textbf{93.6}\,$\pm$\,1.1 \\
\bottomrule
\end{tabularx}
\par\smallskip
\begin{minipage}{\columnwidth}
\scriptsize $^{\dagger}$CRC100K additionally averages over class-name synonyms released with CONCH~\cite{lu2024conch}.
\end{minipage}
\end{table}

%% file: sections/06_conclusion.tex
\section{Conclusion}
\label{sec:conclusion}
We presented \textbf{\methodshort{}}, which aligns separately pretrained vision and language encoders through relative representations defined by \textbf{(1) modality-specific learnable anchors}, with \textbf{(2) anchor-wise similarity pooling} aggregating token evidence separately for each coordinate. Training only the two anchor matrices, \methodshort{} outperforms the compared post-hoc alignment methods on all 13 general-domain benchmarks, and this advantage persists across data scales and encoder combinations.

The ablations show that the two components are complementary. Anchors thus play a dual role, defining the shared relative coordinates and determining which token evidence each coordinate aggregates. The pathology experiments further show that existing specialized image and language models can be connected for zero-shot vision--language tasks without updating their weights, which suggests a practical route to new cross-modal applications of unimodal models.

Several limitations remain. The transfer capabilities of post-hoc alignment depend on the semantic coverage of the paired data, which may be insufficient for concepts requiring detailed texture, scene, or morphological distinctions. The semantic roles of individual anchors also emerge during training rather than being explicitly controlled. These limitations motivate further investigation into improving semantic coverage under limited supervision and regularizing anchors toward more interpretable and controllable roles.

%% file: appendix/title_page.tex
\thispagestyle{empty}
\begingroup
\centering
\vspace*{12pt}
{\huge\bfseries Appendix\par}
\vspace{14pt}
{\Large ReLATE: Learning Relative Representations\par
for Post-Hoc Multimodal Alignment\par}
\vspace{12pt}
{\normalsize Shiwon Kim, JaeSeong Hong, and Yu Rang Park\par}
\vspace{24pt}
\begin{minipage}{0.8\textwidth}
\normalfont\normalsize
\newcommand{\appendixcontentssection}[3]{%
  \par\addvspace{7pt}\noindent
  \textbf{\hyperref[#2]{\makebox[1.8em][l]{#1}#3}}%
  \nobreak\dotfill\hyperref[#2]{\pageref*{#2}}\par}
\newcommand{\appendixcontentssubsection}[2]{%
  \par\noindent\hspace*{1.3em}%
  \hyperref[#2]{\makebox[2.5em][l]{#1}\nameref*{#2}}%
  \nobreak\dotfill\hyperref[#2]{\pageref*{#2}}\par}
\setlength{\parskip}{3pt}
\appendixcontentssection{A}{sec:supp_experimental_setup}{Additional Experimental Settings}
\appendixcontentssubsection{A.1}{app:datasets}
\appendixcontentssubsection{A.2}{app:baselines}
\appendixcontentssubsection{A.3}{app:evaluation_protocols}
\appendixcontentssubsection{A.4}{app:training}
\appendixcontentssection{B}{app:anchor_learning_processing}{Additional Ablation Studies}
\appendixcontentssubsection{B.1}{app:fixed_anchor_processing}
\appendixcontentssubsection{B.2}{app:additional_projection_head}
\appendixcontentssection{C}{sec:supp_data_scaling}{Scaling with Paired Data}
\appendixcontentssubsection{C.1}{app:scaling_110k}
\appendixcontentssubsection{C.2}{app:scaling_datasetwise}
\appendixcontentssection{D}{sec:supp_encoder_pairs}{Dataset-wise Results across Encoder Pairs}
\appendixcontentssection{E}{sec:supp_pathology_prompts}{Additional Pathology Classification Results}
\appendixcontentssubsection{E.1}{app:pathology_training}
\appendixcontentssubsection{E.2}{app:pathology_evaluation}
\appendixcontentssubsection{E.3}{app:pathology_vlms}
\appendixcontentssubsection{E.4}{app:pathology_scaling}
\appendixcontentssubsection{E.5}{app:pathology_results}
\appendixcontentssection{F}{app:anchor_overlap}{Cross-Modal Anchor Consistency}
\end{minipage}
\par
\endgroup
\clearpage

%% file: appendix/content.tex
\renewcommand{\thesubsection}{\thesection.\arabic{subsection}}
\renewcommand{\thesubsectiondis}{\thesubsection}
\renewcommand{\thesubsubsection}{\thesubsection.\arabic{subsubsection}}
\renewcommand{\thesubsubsectiondis}{\thesubsubsection}
\makeatletter
\renewcommand{\p@subsection}{}
\renewcommand{\p@subsubsection}{}
\makeatother
\renewcommand{\theHsubsection}{appendix.\thesection.\arabic{subsection}}
\renewcommand{\theHsubsubsection}{\theHsubsection.\arabic{subsubsection}}
\setcounter{table}{0}
\setcounter{figure}{0}
\renewcommand{\thetable}{A\arabic{table}}
\renewcommand{\thefigure}{A\arabic{figure}}
\renewcommand{\theHtable}{appendix.\arabic{table}}
\renewcommand{\theHfigure}{appendix.\arabic{figure}}
\input{appendix/sections/a_experimental_settings}
\clearpage
\input{appendix/sections/b_ablation_studies}

\clearpage
\input{appendix/sections/c_paired_data_scaling}
\clearpage
\input{appendix/sections/d_encoder_pair_results}

\clearpage
\input{appendix/sections/e_pathology_classification}

\clearpage
\input{appendix/sections/f_cross_modal_anchor_consistency}

%% file: appendix/sections/a_experimental_settings.tex
\begin{appendixtext}
\section{Additional Experimental Settings}
\label{sec:supp_experimental_setup}

\vspace{8pt}
\subsection{Datasets}
\label{app:datasets}
MS COCO is the only dataset used for alignment training; all other datasets below serve as evaluation benchmarks for zero-shot classification, cross-modal retrieval, zero-shot segmentation, and phrase grounding, and none of their evaluation partitions is used for training. The pathology alignment data and benchmarks are described in Appendix~\ref{sec:supp_pathology_prompts}, and the expanded 110K alignment set in Appendix~\ref{sec:supp_data_scaling}.

\vspace{8pt}
\begin{itemize}\setlength{\itemsep}{4pt}
\item \textbf{MS COCO}~\cite{lin2014coco} is a large-scale collection of everyday scenes with object annotations and five human-written captions per image. Alignment uses the 82,783 images of the 2014 training split with the first annotated caption of each image, giving one image--text pair per image.
\item \textbf{STL10}~\cite{coates2011analysis} contains $96\times96$ images of 10 object classes derived from ImageNet. We use the official test split for zero-shot classification.
\item \textbf{CIFAR100}~\cite{krizhevsky2009learning} contains $32\times32$ natural images of 100 fine-grained object classes. We use the official test split of 10,000 images.
\item \textbf{Caltech101}~\cite{fei2004learning} contains object-centric images of 101 categories with varying numbers of images per class. As it provides no official test split, we follow the CLIP protocol~\cite{radford2021learning} and sample the split, holding out up to 30 images per class and evaluating on the remaining images.
\item \textbf{DTD}~\cite{cimpoi2014describing} is a texture dataset of 47 categories named by human-describable attributes, such as \textit{striped} or \textit{bubbly}. We use the first of its ten predefined test partitions.
\item \textbf{EuroSAT}~\cite{helber2019eurosat} consists of Sentinel-2 satellite images covering 10 land-use and land-cover classes. As it provides no official test split, we follow the CLIP protocol~\cite{radford2021learning} and sample the split, drawing 1,500 images per class and using the last 500 of each class for evaluation.
\item \textbf{Flickr30k}~\cite{young2014image} contains images of everyday activities collected from Flickr, each paired with five captions. Retrieval uses the standard test split of 1,000 images and 5,000 captions.
\item \textbf{COCO Karpathy test}~\cite{karpathy2015deep} is the standard 5,000-image retrieval test split drawn from the COCO 2014 validation images, with 25,010 captions. We evaluate on the full 5K gallery. This split serves as a held-out in-domain evaluation.
\item \textbf{PASCAL VOC2012}~\cite{everingham2010pascal} provides pixel-level annotations for 20 object categories. We use the validation split, with background as an additional prediction class and ignore labels excluded. The background class is excluded from the mIoU-fg average.
\item \textbf{PASCAL-Context}~\cite{mottaghi2014role} extends the PASCAL VOC images with whole-scene annotations that include stuff categories such as sky, grass, and wall. We use the validation split with the 59 most frequent classes and background as an additional prediction class. The background class is excluded from the mIoU-fg average.
\item \textbf{ADE20K}~\cite{zhou2017ade20k} is a scene-parsing dataset with dense annotations of diverse indoor and outdoor scenes. We use the validation split of its 150-class benchmark.
\item \textbf{Flickr30k Entities}~\cite{plummer2015entities} augments the Flickr30k captions with bounding boxes for the entity phrases. Phrase grounding uses the phrase--box annotations of the Flickr30k test images.
\item \textbf{ReferItGame}~\cite{kazemzadeh2014referitgame} contains referring expressions for image regions collected through a two-player game on natural images. We use the RefCLEF/Berkeley test split with 65,193 expressions.
\item \textbf{RefCOCOg}~\cite{mao2016generation} contains referring expressions for objects in COCO images, which are longer and more descriptive than those of game-based datasets. We use the UMD test split with 9,602 expressions over 2,600 images. The test images are among the COCO 2014 training images, but referring expressions and box annotations are never used during alignment.
\end{itemize}

\subsection{Baselines}
\label{app:baselines}
We compare against four post-hoc alignment methods: one training-free method (CSA) and three trainable methods (STRUCTURE, SAIL, FreezeAlign). All baselines are reproduced in our setting rather than taken from published numbers. All baselines share the frozen encoders, selected feature layers, alignment pairs, and evaluation splits, and all produce 512-dimensional representations. Baseline-specific hyperparameters follow the official implementations unless stated otherwise.

\vspace{8pt}
\begin{itemize}\setlength{\itemsep}{4pt}
\item \textbf{CSA} (ICLR 2025)~\cite{csa2025} is a training-free method that fits canonical correlation analysis on pooled image and text features of paired samples and compares the projected features with a similarity weighted by the canonical correlations. We set the number of canonical dimensions to 512 to match the other methods.
\item \textbf{STRUCTURE} (NeurIPS 2025)~\cite{structure2025} trains lightweight alignment heads for limited paired data, regularizing them to preserve the multiscale neighborhood structure of each unimodal space and selecting encoder layers by representational similarity. Of its linear and MLP variants, we use the MLP variant, which follows the official implementation with a rank-64 residual head and a learnable gate and takes CLS image features and mean-pooled token features as input.
\item \textbf{SAIL} (CVPR 2025)~\cite{sail2025} trains a nonlinear gated alignment layer on top of frozen vision and language models. Following its DINOv2 configuration, the image branch concatenates the CLS feature with the mean patch feature before the gated StarMLP mapping, and the text branch maps mean-pooled token features.
\item \textbf{FreezeAlign} (CVPR 2025)~\cite{freezealign} trains projection modules on frozen unimodal encoders using both global and patch-level features. Its image branch projects patch and CLS features separately and adds the averaged projected patches to the projected CLS feature, and its text branch applies a residual MLP head to masked-mean-pooled token features.
\end{itemize}
\vspace{8pt}

Table~\ref{tab:alignment_parameters} compares the trainable alignment parameters of all baselines for the default encoder pair, excluding the frozen encoders. \methodshort{} learns $K(D_v+D_t)=1{,}048{,}576$ parameters at $K{=}512$, compared with 1.2M for STRUCTURE, 4.5M for FreezeAlign, and 96.5M for SAIL; CSA has no gradient-trained parameters (CCA transformations are fitted from the paired data).
\end{appendixtext}

\input{appendix/tables/a01_alignment_parameter_counts}

\begin{appendixtext}

\subsection{Evaluation Protocols}
\label{app:evaluation_protocols}
After alignment, the encoders and alignment parameters are fixed, and no task-specific training, labels, or postprocessing are used. The same protocol and evaluation code are applied to every method.

Classification inserts each class name into dataset-specific templates, normalizes each prompt representation, averages them into a class prototype, and renormalizes. Images are assigned to the prototype with the highest cosine similarity, and top-1 accuracy is the fraction of correctly classified images. The templates for each dataset are listed below. The placeholder \texttt{\{class\}} is replaced by the class name, and the wording is retained as used in evaluation. Histopathology templates and CRC100K class-name synonyms are listed in Appendix~\ref{sec:supp_pathology_prompts}.

\vspace{8pt}
\begin{itemize}\setlength{\itemsep}{4pt}
\item \textbf{STL10} (2 templates).\par
{\footnotesize\raggedright \texttt{a photo of a \{class\}.}
\quad \texttt{a photo of the \{class\}.}\par}
\item \textbf{CIFAR100} (18 templates).\par
{\footnotesize\raggedright \texttt{a photo of a \{class\}.}
\quad \texttt{a blurry photo of a \{class\}.}
\quad \texttt{a black and white photo of a \{class\}.}
\quad \texttt{a low contrast photo of a \{class\}.}
\quad \texttt{a high contrast photo of a \{class\}.}
\quad \texttt{a bad photo of a \{class\}.}
\quad \texttt{a good photo of a \{class\}.}
\quad \texttt{a photo of a small \{class\}.}
\quad \texttt{a photo of a big \{class\}.}
\quad \texttt{a photo of the \{class\}.}
\quad \texttt{a blurry photo of the \{class\}.}
\quad \texttt{a black and white photo of the \{class\}.}
\quad \texttt{a low contrast photo of the \{class\}.}
\quad \texttt{a high contrast photo of the \{class\}.}
\quad \texttt{a bad photo of the \{class\}.}
\quad \texttt{a good photo of the \{class\}.}
\quad \texttt{a photo of the small \{class\}.}
\quad \texttt{a photo of the big \{class\}.}\par}
\item \textbf{Caltech101} (34 templates).\par
{\footnotesize\raggedright \texttt{a photo of a \{class\}.}
\quad \texttt{a painting of a \{class\}.}
\quad \texttt{a plastic \{class\}.}
\quad \texttt{a sculpture of a \{class\}.}
\quad \texttt{a sketch of a \{class\}.}
\quad \texttt{a tattoo of a \{class\}.}
\quad \texttt{a toy \{class\}.}
\quad \texttt{a rendition of a \{class\}.}
\quad \texttt{a embroidered \{class\}.}
\quad \texttt{a cartoon \{class\}.}
\quad \texttt{a \{class\} in a video game.}
\quad \texttt{a plushie \{class\}.}
\quad \texttt{a origami \{class\}.}
\quad \texttt{art of a \{class\}.}
\quad \texttt{graffiti of a \{class\}.}
\quad \texttt{a drawing of a \{class\}.}
\quad \texttt{a doodle of a \{class\}.}
\quad \texttt{a photo of the \{class\}.}
\quad \texttt{a painting of the \{class\}.}
\quad \texttt{the plastic \{class\}.}
\quad \texttt{a sculpture of the \{class\}.}
\quad \texttt{a sketch of the \{class\}.}
\quad \texttt{a tattoo of the \{class\}.}
\quad \texttt{the toy \{class\}.}
\quad \texttt{a rendition of the \{class\}.}
\quad \texttt{the embroidered \{class\}.}
\quad \texttt{the cartoon \{class\}.}
\quad \texttt{the \{class\} in a video game.}
\quad \texttt{the plushie \{class\}.}
\quad \texttt{the origami \{class\}.}
\quad \texttt{art of the \{class\}.}
\quad \texttt{graffiti of the \{class\}.}
\quad \texttt{a drawing of the \{class\}.}
\quad \texttt{a doodle of the \{class\}.}\par}
\item \textbf{DTD} (8 templates).\par
{\footnotesize\raggedright \texttt{a photo of a \{class\} texture.}
\quad \texttt{a photo of a \{class\} pattern.}
\quad \texttt{a photo of a \{class\} thing.}
\quad \texttt{a photo of a \{class\} object.}
\quad \texttt{a photo of the \{class\} texture.}
\quad \texttt{a photo of the \{class\} pattern.}
\quad \texttt{a photo of the \{class\} thing.}
\quad \texttt{a photo of the \{class\} object.}\par}
\item \textbf{EuroSAT} (3 templates).\par
{\footnotesize\raggedright \texttt{a centered satellite photo of \{class\}.}
\quad \texttt{a centered satellite photo of a \{class\}.}
\quad \texttt{a centered satellite photo of the \{class\}.}\par}
\end{itemize}
\vspace{12pt}

Retrieval ranks image and text candidates by cosine similarity. Image-to-text recall@1 counts a hit when the top-ranked caption belongs to the query image, and text-to-image recall@1 uses every caption as a query against the unique images. Each image contributes once per associated caption, so Flickr30k weights its images equally while COCO follows the caption counts of the annotations.

Segmentation forms class prototypes from the 80 ImageNet templates used by CLIP~\cite{radford2021learning}, computes cosine similarities between normalized patch descriptors and prototypes, bilinearly upsamples the score maps to the mask resolution, and assigns each pixel to the highest-scoring class without postprocessing. IoU is computed per class from a dataset-level confusion matrix, and mIoU-fg averages the foreground classes. The 80 templates are listed below.

\vspace{8pt}
\begin{itemize}\setlength{\itemsep}{4pt}
\item \textbf{ImageNet} (80 templates).\par
{\footnotesize\raggedright \texttt{a bad photo of a \{class\}.}
\quad \texttt{a photo of many \{class\}.}
\quad \texttt{a sculpture of a \{class\}.}
\quad \texttt{a photo of the hard to see \{class\}.}
\quad \texttt{a low resolution photo of the \{class\}.}
\quad \texttt{a rendering of a \{class\}.}
\quad \texttt{graffiti of a \{class\}.}
\quad \texttt{a bad photo of the \{class\}.}
\quad \texttt{a cropped photo of the \{class\}.}
\quad \texttt{a tattoo of a \{class\}.}
\quad \texttt{the embroidered \{class\}.}
\quad \texttt{a photo of a hard to see \{class\}.}
\quad \texttt{a bright photo of a \{class\}.}
\quad \texttt{a photo of a clean \{class\}.}
\quad \texttt{a photo of a dirty \{class\}.}
\quad \texttt{a dark photo of the \{class\}.}
\quad \texttt{a drawing of a \{class\}.}
\quad \texttt{a photo of my \{class\}.}
\quad \texttt{the plastic \{class\}.}
\quad \texttt{a photo of the cool \{class\}.}
\quad \texttt{a close-up photo of a \{class\}.}
\quad \texttt{a black and white photo of the \{class\}.}
\quad \texttt{a painting of the \{class\}.}
\quad \texttt{a painting of a \{class\}.}
\quad \texttt{a pixelated photo of the \{class\}.}
\quad \texttt{a sculpture of the \{class\}.}
\quad \texttt{a bright photo of the \{class\}.}
\quad \texttt{a cropped photo of a \{class\}.}
\quad \texttt{a plastic \{class\}.}
\quad \texttt{a photo of the dirty \{class\}.}
\quad \texttt{a jpeg corrupted photo of a \{class\}.}
\quad \texttt{a blurry photo of the \{class\}.}
\quad \texttt{a photo of the \{class\}.}
\quad \texttt{a good photo of the \{class\}.}
\quad \texttt{a rendering of the \{class\}.}
\quad \texttt{a \{class\} in a video game.}
\quad \texttt{a photo of one \{class\}.}
\quad \texttt{a doodle of a \{class\}.}
\quad \texttt{a close-up photo of the \{class\}.}
\quad \texttt{a photo of a \{class\}.}
\quad \texttt{the origami \{class\}.}
\quad \texttt{the \{class\} in a video game.}
\quad \texttt{a sketch of a \{class\}.}
\quad \texttt{a doodle of the \{class\}.}
\quad \texttt{a origami \{class\}.}
\quad \texttt{a low resolution photo of a \{class\}.}
\quad \texttt{the toy \{class\}.}
\quad \texttt{a rendition of the \{class\}.}
\quad \texttt{a photo of the clean \{class\}.}
\quad \texttt{a photo of a large \{class\}.}
\quad \texttt{a rendition of a \{class\}.}
\quad \texttt{a photo of a nice \{class\}.}
\quad \texttt{a photo of a weird \{class\}.}
\quad \texttt{a blurry photo of a \{class\}.}
\quad \texttt{a cartoon \{class\}.}
\quad \texttt{art of a \{class\}.}
\quad \texttt{a sketch of the \{class\}.}
\quad \texttt{a embroidered \{class\}.}
\quad \texttt{a pixelated photo of a \{class\}.}
\quad \texttt{itap of the \{class\}.}
\quad \texttt{a jpeg corrupted photo of the \{class\}.}
\quad \texttt{a good photo of a \{class\}.}
\quad \texttt{a plushie \{class\}.}
\quad \texttt{a photo of the nice \{class\}.}
\quad \texttt{a photo of the small \{class\}.}
\quad \texttt{a photo of the weird \{class\}.}
\quad \texttt{the cartoon \{class\}.}
\quad \texttt{art of the \{class\}.}
\quad \texttt{a drawing of the \{class\}.}
\quad \texttt{a photo of the large \{class\}.}
\quad \texttt{a black and white photo of a \{class\}.}
\quad \texttt{the plushie \{class\}.}
\quad \texttt{a dark photo of a \{class\}.}
\quad \texttt{itap of a \{class\}.}
\quad \texttt{graffiti of the \{class\}.}
\quad \texttt{a toy \{class\}.}
\quad \texttt{itap of my \{class\}.}
\quad \texttt{a photo of a cool \{class\}.}
\quad \texttt{a photo of a small \{class\}.}
\quad \texttt{a tattoo of the \{class\}.}\par}
\end{itemize}
\vspace{12pt}

Phrase grounding uses the annotated phrases as queries without templates. Patch--phrase cosine similarities are upsampled to the image resolution, and the highest-scoring pixel is the predicted pointing location. Pointing accuracy~\cite{akbari2019multilevel} is the percentage of queries whose predicted location falls inside a ground-truth box; for Flickr30k phrases with multiple boxes, a hit in any box counts, and phrases without boxes are excluded.

For segmentation and grounding, each method keeps spatial patch descriptors instead of pooling them: STRUCTURE maps each patch through its alignment head, SAIL concatenates each patch with the image CLS feature before mapping, FreezeAlign uses its local visual projection, CSA applies its fitted transformation to patches, and \methodshort{} uses patch-wise anchor similarities. Descriptors and class or phrase representations are normalized before cosine similarity, and score-map interpolation and evaluation rules are shared across methods.

Multi-seed experiments use seeds 42, 44, and 123. For trainable methods, each seed defines one training run. CSA is deterministic, so it is fitted once on each full paired set. At smaller data sizes, the same three seeds draw the paired subsets for all methods. Task averages are computed within each seed before averaging across seeds.

\subsection{Training and Hyperparameters}
\label{app:training}
The settings below apply to all general-domain experiments with the default encoder pair. Settings that differ for other encoder pairs and for pathology are given in Appendices~\ref{sec:supp_encoder_pairs} and~\ref{sec:supp_pathology_prompts}, respectively.

The default encoders are DINOv2 ViT-L/14~\cite{oquab2023dinov2} and RoBERTa-Large~\cite{liu2019roberta}, both with feature dimension 1,024. Following STRUCTURE~\cite{structure2025}, we select the image block and text layer with the largest mutual-kNN similarity on 5,000 sampled alignment pairs, giving image block 23 and text layer 24. Images use resolution $224\times224$; segmentation and grounding resize the whole image without center cropping. The image sequence contains a CLS token and 256 spatial patches, and dense prediction uses only the spatial patches.

\methodshort{} initializes its two anchor matrices from normal random values with row normalization and uses $K{=}512$, pooling temperature $\tau_p{=}0.03$, and contrastive temperature $\tau{=}0.05$; both temperatures are fixed. Trainable methods use AdamW~\cite{loshchilov2019decoupled} with batch size 4,096, betas $(0.9,0.95)$, weight decay $10^{-4}$, and gradient-norm clipping at 1.0. The learning rate of each run is chosen by a range test over logarithmically increasing rates and then follows cosine annealing. All alignment training uses a single NVIDIA Quadro RTX 8000 GPU (48\,GB). All hyperparameters, including $\tau_p$, are shared across benchmarks and were not tuned per dataset.
\end{appendixtext}

%% file: appendix/tables/a01_alignment_parameter_counts.tex
\begin{table}[!ht]
\centering
\begin{minipage}{0.6\textwidth}
\caption{Alignment parameter counts for the DINOv2 ViT-L and RoBERTa-Large encoder pair. Counts exclude frozen encoders. Values are reported in millions, rounded to one decimal place.}
\label{tab:alignment_parameters}
\end{minipage}
\centering
\footnotesize
\renewcommand{\arraystretch}{1.20}
\setlength{\tabcolsep}{9pt}
\begin{tabular}{lcc}
\toprule
\textbf{Method} & \textbf{Output Dimension} & \textbf{Params. (M)} \\
\midrule
CSA~\cite{csa2025} & 512 & 0 \\
STRUCTURE~\cite{structure2025} & 512 & 1.2 \\
SAIL~\cite{sail2025} & 512 & 96.5 \\
FreezeAlign~\cite{freezealign} & 512 & 4.5 \\
\midrule
\rowcolor{relatehighlight}
\textbf{\methodshort{} (ours)} & 512 & 1.0 \\
\bottomrule
\end{tabular}
\end{table}

%% file: appendix/sections/b_ablation_studies.tex
\begin{appendixtext}
\section{Additional Ablation Studies}
\label{app:anchor_learning_processing}

\vspace{8pt}
\subsection{Similarity Processing for Fixed Anchors}
\label{app:fixed_anchor_processing}
The fixed anchor variants in Table~III of the main paper replace the learnable anchors of \methodshort{} with frozen paired-sample anchors. Table~\ref{tab:anchor_learning_processing} examines how these variants are constructed at $K{=}512$, with and without the ASIF-style similarity processing~\cite{norelli2023asif}. Fixed anchors are the pooled embeddings of corresponding image--caption pairs and remain frozen; token-level mean, max, and ASP pooling then act on the similarities to these anchors, with $\tau_p{=}0.03$ for ASP. The ASIF-style processing retains the top-$k$ anchor similarities (all 512 at $K{=}512$ and the top 800 at $K{=}82$K), clamps them to nonnegative values, raises them to the eighth power, and $\ell_2$-normalizes the result. This processing improves fixed anchors on classification, retrieval, and segmentation. ASP does not improve on mean pooling in either setting, and the fixed anchor variants remain far below \methodshort{} on all four tasks.
\end{appendixtext}

\input{appendix/tables/a02_fixed_anchor_similarity_processing}

\begin{appendixtext}
\subsection{Additional Projection Head}
\label{app:additional_projection_head}
We evaluate whether adding an explicit projection head improves \methodshort{}. Modality-specific 512-dimensional bottleneck projectors are inserted before the anchor-based module, while both variants use the same paired training set and the ViT-L / RoBERTa-Large encoder pair. Table~\ref{tab:additional_projection_head} reports a single-seed comparison (seed 42). Adding the projectors reduces performance on all four evaluated tasks. Figure~\ref{fig:projector_loss_curves} shows the training and validation losses over 200 epochs; the additional projectors yield lower training loss but higher validation loss, increasing the train--validation gap. This comparison provides no evidence of a benefit from additional projection heads under this experimental setting.
\end{appendixtext}

\input{appendix/tables/a03_projection_head_ablation}

%% file: appendix/tables/a02_fixed_anchor_similarity_processing.tex
\begin{table}[!ht]
\centering
\begin{minipage}{0.8\textwidth}
\caption{Fixed $K{=}512$ anchors with and without the ASIF-style similarity processing~\cite{norelli2023asif}. Scores are task averages (\%), reported as mean $\pm$ standard deviation over three seeds. \methodshort{} with $K{=}512$ learnable anchors is shown for reference.}
\label{tab:anchor_learning_processing}
\centering
\renewcommand{\arraystretch}{1.20}
\footnotesize
\setlength{\tabcolsep}{3pt}
\begin{tabularx}{\linewidth}{l >{\centering\arraybackslash}p{94pt} l *{4}{>{\centering\arraybackslash}X}}
\toprule
\textbf{Anchors} & \textbf{Similarity Processing} & \textbf{Pooling} & \textbf{Avg. Cls.} & \textbf{Avg. Ret.} & \textbf{Avg. Seg.} & \textbf{Avg. Gr.} \\
\midrule
\multirow{3}{*}{Fixed} & \multirow{3}{*}{\xmark} & Mean & 9.8\,$\pm$\,0.9 & 1.8\,$\pm$\,0.1 & 7.4\,$\pm$\,0.1 & 40.1\,$\pm$\,0.5 \\
 & & Max & 6.2\,$\pm$\,0.3 & 1.7\,$\pm$\,0.1 & 7.8\,$\pm$\,0.2 & 37.3\,$\pm$\,0.3 \\
 & & ASP & 6.6\,$\pm$\,0.5 & 1.7\,$\pm$\,0.1 & 7.6\,$\pm$\,0.2 & 38.3\,$\pm$\,0.4 \\
\midrule
\multirow{3}{*}{Fixed~\cite{norelli2023asif}} & \multirow{3}{*}{\cmark} & Mean & 29.0\,$\pm$\,1.8 & 3.9\,$\pm$\,0.3 & 11.9\,$\pm$\,0.6 & 39.4\,$\pm$\,0.5 \\
 & & Max & 23.6\,$\pm$\,1.5 & 3.1\,$\pm$\,0.3 & 11.6\,$\pm$\,0.8 & 39.2\,$\pm$\,0.5 \\
 & & ASP & 23.9\,$\pm$\,1.8 & 2.9\,$\pm$\,0.4 & 11.5\,$\pm$\,0.9 & 39.3\,$\pm$\,0.5 \\
\midrule
Learnable & \xmark & \cellcolor{relatehighlight}\textbf{ASP (\methodshort{})} & \cellcolor{relatehighlight}\textbf{51.2}\,$\pm$\,0.3 & \cellcolor{relatehighlight}\textbf{59.2}\,$\pm$\,0.5 & \cellcolor{relatehighlight}\textbf{23.9}\,$\pm$\,0.3 & \cellcolor{relatehighlight}\textbf{59.0}\,$\pm$\,0.3 \\
\bottomrule
\end{tabularx}
\end{minipage}
\end{table}

%% file: appendix/tables/a03_projection_head_ablation.tex
\begin{table}[!ht]
\centering
\begin{minipage}[t]{0.47\textwidth}
\vspace{0pt}
\caption{Effect of adding bottleneck projectors before the anchor-based module. Both variants use the same paired training set and seed 42. Scores are task averages (\%).}
\label{tab:additional_projection_head}
\centering
\renewcommand{\arraystretch}{1.20}
\footnotesize
\setlength{\tabcolsep}{2pt}
\begin{tabularx}{\linewidth}{l *{4}{>{\centering\arraybackslash}X}}
\toprule
\textbf{Method} & \textbf{Avg. Cls.} & \textbf{Avg. Ret.} & \textbf{Avg. Seg.} & \textbf{Avg. Gr.} \\
\midrule
\rowcolor{relatehighlight}
\textbf{\methodshort{}} & \textbf{51.5} & \textbf{59.3} & \textbf{23.8} & \textbf{58.7} \\
\methodshort{} + Proj. & 37.3 & 37.0 & 2.1 & 42.4 \\
\bottomrule
\end{tabularx}
\end{minipage}\hfill
\begin{minipage}[t]{0.49\textwidth}
\vspace{0pt}
\makeatletter\def\@captype{figure}\makeatother
\centering
\includegraphics[width=\linewidth]{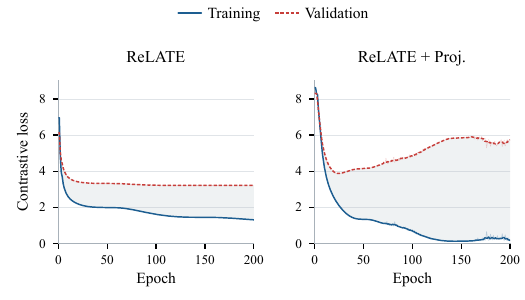}
\caption{Training and validation contrastive losses with and without additional projectors. Thin lines show raw losses, and prominent lines show five-epoch moving averages. Shading indicates the gap between the smoothed curves.}
\label{fig:projector_loss_curves}
\end{minipage}
\end{table}

%% file: appendix/sections/c_paired_data_scaling.tex
\begin{appendixtext}
\section{Scaling with Paired Data}
\label{sec:supp_data_scaling}

\vspace{8pt}
\subsection{Scaling to 110K Paired Samples}
\label{app:scaling_110k}
Table~\ref{tab:supp_data_scaling_110k} compares task-averaged performance with the 82K COCO 2014 training pairs and the expanded 110K alignment set, including the change for each method and task. The expanded set is built from COCO 2017 training annotations by removing the images of the COCO Karpathy test split and keeping the first caption per image, giving 113,880 pairs (denoted 110K). CSA is evaluated once; the trainable methods use three seeds. Increasing the alignment set from 82K to 110K improves \methodshort{} on all four tasks, while it maintains higher performance than the compared baselines.
\end{appendixtext}

\input{appendix/tables/a04_paired_data_scaling_82k_110k}

\begin{figure}[!ht]
\centering
\includegraphics[width=\textwidth]{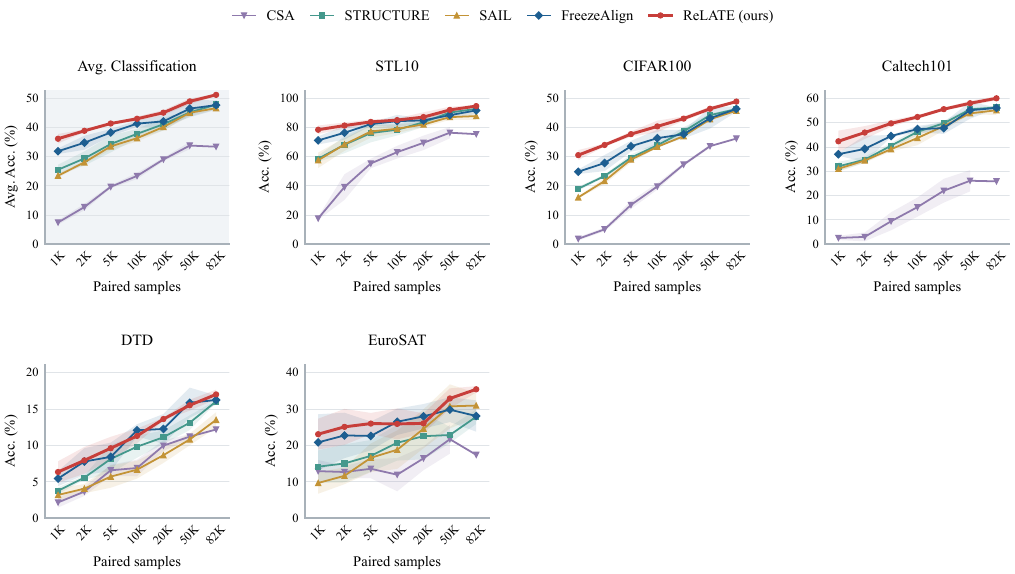}
\caption{Dataset-wise zero-shot classification as the paired alignment set increases from 1K to 82K. The first panel shows the task average. Lines show mean accuracy (\%); shaded bands indicate standard deviation over three runs. CSA uses three sampled subsets at smaller sizes and a single full-data evaluation.}
\label{fig:supp_scaling_classification}
\end{figure}

\begin{figure}[!ht]
\centering
\includegraphics[width=\textwidth]{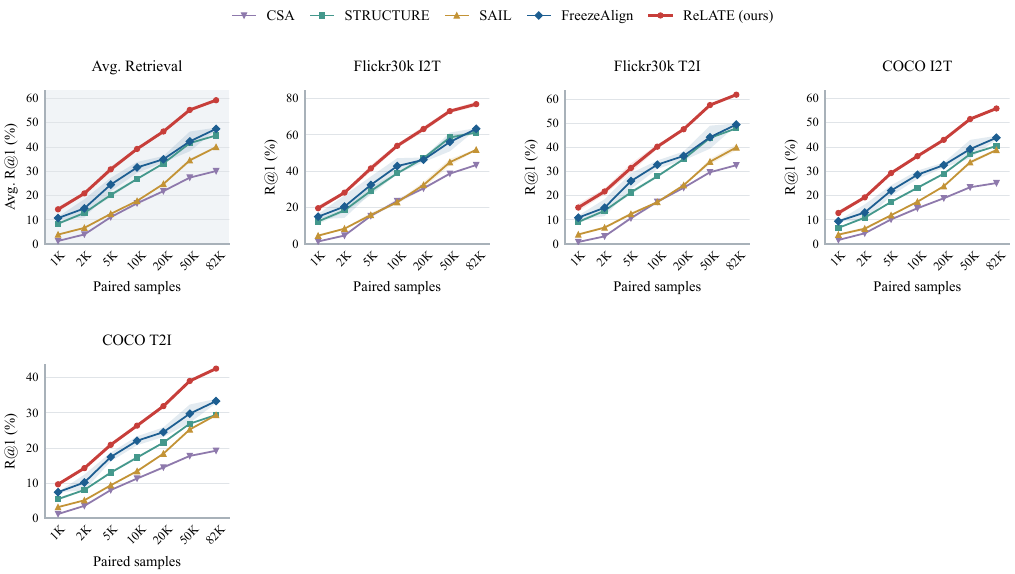}
\caption{Dataset-wise cross-modal retrieval as the paired alignment set increases from 1K to 82K. The first panel shows the task average. Lines show mean R@1 (\%); shaded bands indicate standard deviation over three runs. CSA uses three sampled subsets at smaller sizes and a single full-data evaluation.}
\label{fig:supp_scaling_retrieval}
\end{figure}

\begin{figure}[!ht]
\centering
\includegraphics[width=\textwidth]{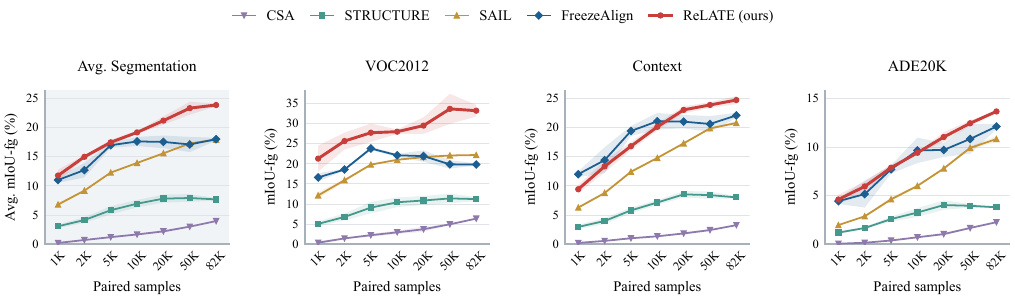}
\caption{Dataset-wise zero-shot segmentation as the paired alignment set increases from 1K to 82K. The first panel shows the task average. Lines show mean mIoU-fg (\%); shaded bands indicate standard deviation over three runs. CSA uses three sampled subsets at smaller sizes and a single full-data evaluation.}
\label{fig:supp_scaling_segmentation}
\end{figure}

\begin{figure}[!ht]
\centering
\includegraphics[width=\textwidth]{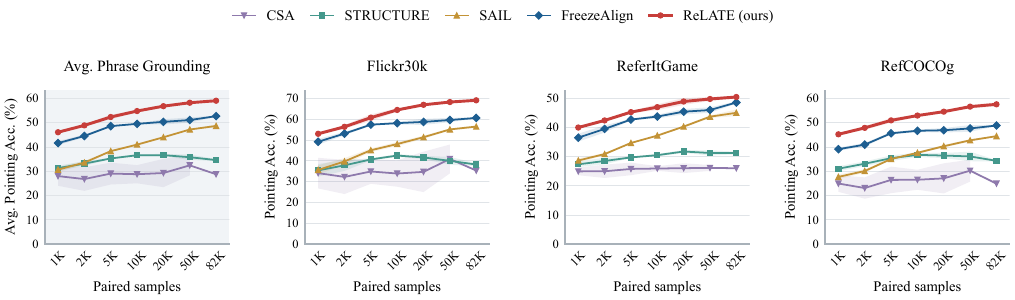}
\caption{Dataset-wise phrase grounding as the paired alignment set increases from 1K to 82K. The first panel shows the task average. Lines show mean pointing acc. (\%); shaded bands indicate standard deviation over three runs. CSA uses three sampled subsets at smaller sizes and a single full-data evaluation.}
\label{fig:supp_scaling_grounding}
\end{figure}

\begin{appendixtext}
\subsection{Dataset-wise Scaling from 1K to Full}
\label{app:scaling_datasetwise}
Figures~\ref{fig:supp_scaling_classification}--\ref{fig:supp_scaling_grounding} give the dataset-wise data scaling curves. Subsets of 1K, 2K, 5K, 10K, 20K, and 50K pairs are sampled from the full alignment pool, with the validation set and training settings of Section~\ref{sec:supp_experimental_setup} unchanged and the learning rate selected per run. Shaded bands show standard deviation over three seeds; CSA uses three sampled subsets at smaller sizes and a single full-data evaluation. Retrieval directions are shown separately, and grounding uses test splits only.
\end{appendixtext}

\clearpage

\clearpage

%% file: appendix/tables/a04_paired_data_scaling_82k_110k.tex
\begin{table}[!ht]
\centering
\begin{minipage}{0.8\textwidth}
\caption{Task-averaged performance with 82K and 110K paired alignment samples. Scores are percentages; trainable methods report mean $\pm$ standard deviation over three seeds, and CSA uses a single evaluation. Parentheses show changes from 82K in percentage points, with increases in green and decreases in red. Best and second-best results within each data budget are in \textbf{bold} and \underline{underlined}.}
\label{tab:supp_data_scaling_110k}
\centering
\footnotesize
\renewcommand{\arraystretch}{1.20}
\setlength{\tabcolsep}{2pt}
\newlength{\scalingentrywidth}
\settowidth{\scalingentrywidth}{\makebox[15pt][l]{\textbf{99.9}}\makebox[23pt][l]{\,$\pm$\,9.9}\,\textbf{(+9.9)}}
\newcommand{\scalingentry}[3]{\makebox[15pt][l]{#1}\makebox[23pt][l]{#2}#3}
\begin{tabularx}{\linewidth}{l >{\centering\arraybackslash}p{52pt} *{4}{>{\raggedright\arraybackslash\hspace*{\dimexpr(\hsize-\scalingentrywidth)/2\relax}}X}}
\toprule
\textbf{Method} & \textbf{Paired Data} & \multicolumn{1}{c}{\textbf{Avg. Cls.}} & \multicolumn{1}{c}{\textbf{Avg. Ret.}} & \multicolumn{1}{c}{\textbf{Avg. Seg.}} & \multicolumn{1}{c}{\textbf{Avg. Gr.}} \\
\midrule
CSA~\cite{csa2025} & 82K & \scalingentry{33.4}{}{} & \scalingentry{30.1}{}{} & \scalingentry{3.9}{}{} & \scalingentry{28.7}{}{} \\
 & 110K & \scalingentry{35.3}{}{\,\textcolor[rgb]{0,0.45,0}{\textbf{(+1.9)}}} & \scalingentry{31.7}{}{\,\textcolor[rgb]{0,0.45,0}{\textbf{(+1.6)}}} & \scalingentry{4.1}{}{\,\textcolor[rgb]{0,0.45,0}{\textbf{(+0.2)}}} & \scalingentry{24.8}{}{\,\textcolor[rgb]{0.75,0.12,0.12}{\textbf{(-3.9)}}} \\
\midrule
STRUCTURE~\cite{structure2025} & 82K & \scalingentry{\underline{47.9}}{\,$\pm$\,0.2}{} & \scalingentry{44.7}{\,$\pm$\,0.2}{} & \scalingentry{7.7}{\,$\pm$\,0.2}{} & \scalingentry{34.6}{\,$\pm$\,0.4}{} \\
 & 110K & \scalingentry{\underline{50.3}}{\,$\pm$\,0.3}{\,\textcolor[rgb]{0,0.45,0}{\textbf{(+2.4)}}} & \scalingentry{47.0}{\,$\pm$\,0.3}{\,\textcolor[rgb]{0,0.45,0}{\textbf{(+2.3)}}} & \scalingentry{7.1}{\,$\pm$\,0.2}{\,\textcolor[rgb]{0.75,0.12,0.12}{\textbf{(-0.6)}}} & \scalingentry{32.8}{\,$\pm$\,0.2}{\,\textcolor[rgb]{0.75,0.12,0.12}{\textbf{(-1.8)}}} \\
\midrule
SAIL~\cite{sail2025} & 82K & \scalingentry{46.6}{\,$\pm$\,0.4}{} & \scalingentry{40.0}{\,$\pm$\,0.2}{} & \scalingentry{17.9}{\,$\pm$\,0.2}{} & \scalingentry{48.6}{\,$\pm$\,0.6}{} \\
 & 110K & \scalingentry{48.2}{\,$\pm$\,0.7}{\,\textcolor[rgb]{0,0.45,0}{\textbf{(+1.6)}}} & \scalingentry{42.9}{\,$\pm$\,0.5}{\,\textcolor[rgb]{0,0.45,0}{\textbf{(+2.9)}}} & \scalingentry{\underline{18.3}}{\,$\pm$\,0.2}{\,\textcolor[rgb]{0,0.45,0}{\textbf{(+0.4)}}} & \scalingentry{49.5}{\,$\pm$\,0.5}{\,\textcolor[rgb]{0,0.45,0}{\textbf{(+0.9)}}} \\
\midrule
FreezeAlign~\cite{freezealign} & 82K & \scalingentry{47.7}{\,$\pm$\,1.6}{} & \scalingentry{\underline{47.5}}{\,$\pm$\,0.5}{} & \scalingentry{\underline{18.0}}{\,$\pm$\,0.2}{} & \scalingentry{\underline{52.7}}{\,$\pm$\,0.2}{} \\
 & 110K & \scalingentry{49.3}{\,$\pm$\,0.0}{\,\textcolor[rgb]{0,0.45,0}{\textbf{(+1.6)}}} & \scalingentry{\underline{47.2}}{\,$\pm$\,1.3}{\,\textcolor[rgb]{0.75,0.12,0.12}{\textbf{(-0.3)}}} & \scalingentry{16.6}{\,$\pm$\,0.5}{\,\textcolor[rgb]{0.75,0.12,0.12}{\textbf{(-1.4)}}} & \scalingentry{\underline{53.3}}{\,$\pm$\,0.3}{\,\textcolor[rgb]{0,0.45,0}{\textbf{(+0.6)}}} \\
\midrule
\rowcolor{relatehighlight}
\textbf{\methodshort{} (ours)} & 82K & \scalingentry{\textbf{51.2}}{\,$\pm$\,0.3}{} & \scalingentry{\textbf{59.2}}{\,$\pm$\,0.5}{} & \scalingentry{\textbf{23.9}}{\,$\pm$\,0.3}{} & \scalingentry{\textbf{59.0}}{\,$\pm$\,0.3}{} \\
\rowcolor{relatehighlight}
 & 110K & \scalingentry{\textbf{52.2}}{\,$\pm$\,0.9}{\,\textcolor[rgb]{0,0.45,0}{\textbf{(+1.0)}}} & \scalingentry{\textbf{61.7}}{\,$\pm$\,0.4}{\,\textcolor[rgb]{0,0.45,0}{\textbf{(+2.5)}}} & \scalingentry{\textbf{24.7}}{\,$\pm$\,0.9}{\,\textcolor[rgb]{0,0.45,0}{\textbf{(+0.8)}}} & \scalingentry{\textbf{60.1}}{\,$\pm$\,0.2}{\,\textcolor[rgb]{0,0.45,0}{\textbf{(+1.1)}}} \\
\bottomrule
\end{tabularx}
\end{minipage}
\end{table}

%% file: appendix/sections/d_encoder_pair_results.tex
\begin{appendixtext}
\section{Dataset-wise Results across Encoder Pairs}
\label{sec:supp_encoder_pairs}

\vspace{8pt}
Tables~\ref{tab:supp_encoder_classification}--\ref{tab:supp_encoder_grounding} report the dataset-wise results for all six encoder pairs and five methods, each from a single seed (seed 42 for the default pair). All vision encoders are DINOv2 (ViT-L/14 and ViT-B/14); the text encoders are RoBERTa-Large (all-roberta-large-v1), MPNet (all-mpnet-base-v2), Qwen2.5-0.5B, and gte-Qwen2-1.5B-instruct. The selected image/text layers are $(23,24)$ for ViT-L/RoBERTa, $(23,12)$ for ViT-L/MPNet, $(23,23)$ for ViT-L/Qwen2.5, $(23,28)$ for ViT-L/gte-Qwen2, $(11,24)$ for ViT-B/RoBERTa, and $(11,12)$ for ViT-B/MPNet, all at $224\times224$ resolution. Where a baseline requires a sample-level text feature, RoBERTa and MPNet use mean pooling and Qwen2.5 and gte-Qwen2 use last-token pooling; \methodshort{} applies ASP to the token sequence.
\end{appendixtext}

\input{appendix/tables/a05_encoder_pairs_classification}

\clearpage

\input{appendix/tables/a06_encoder_pairs_retrieval}

\clearpage

\input{appendix/tables/a07_encoder_pairs_segmentation}

\clearpage

\input{appendix/tables/a08_encoder_pairs_grounding}

%% file: appendix/tables/a05_encoder_pairs_classification.tex
\begin{table}[!ht]
\caption{Zero-shot classification across encoder pairs (accuracy, \%). Best results within each encoder pair and column are in \textbf{bold}; second-best results are \underline{underlined}.}
\label{tab:supp_encoder_classification}
\centering
\footnotesize
\renewcommand{\arraystretch}{1.20}
\setlength{\tabcolsep}{3pt}
\begin{tabularx}{\textwidth}{p{0.13\textwidth} p{0.14\textwidth} l *{6}{>{\centering\arraybackslash}X}}
\toprule
\textbf{Vision Model} & \textbf{Language Model} & \textbf{Method} & STL10 & CIFAR100 & Caltech101 & DTD & EuroSAT & \textbf{Avg.} \\
\midrule
\multirow{20}{*}{DINOv2 ViT-L} & \multirow{5}{*}{RoBERTa-Large} & CSA~\cite{csa2025} & 75.5 & 36.2 & 25.8 & 12.2 & 17.4 & 33.4 \\
 &  & STRUCTURE~\cite{structure2025} & 92.8 & \underline{46.6} & 56.9 & 16.2 & 25.7 & 47.6 \\
 &  & SAIL~\cite{sail2025} & 88.8 & 45.8 & \underline{57.1} & 13.6 & 27.3 & 46.5 \\
 &  & FreezeAlign~\cite{freezealign} & \underline{93.4} & 45.9 & 56.3 & \underline{16.6} & \underline{30.4} & \underline{48.5} \\
 &  & \cellcolor{relatehighlight}\textbf{\methodshort{} (ours)} & \cellcolor{relatehighlight}\textbf{95.3} & \cellcolor{relatehighlight}\textbf{48.8} & \cellcolor{relatehighlight}\textbf{60.9} & \cellcolor{relatehighlight}\textbf{17.7} & \cellcolor{relatehighlight}\textbf{34.6} & \cellcolor{relatehighlight}\textbf{51.5} \\
\cmidrule(lr){2-9}
 & \multirow{5}{*}{Qwen2.5-0.5B} & CSA~\cite{csa2025} & 75.5 & 29.0 & 16.8 & 9.3 & 20.7 & 30.3 \\
 &  & STRUCTURE~\cite{structure2025} & 78.5 & 38.9 & 45.2 & \underline{13.6} & \underline{31.4} & 41.5 \\
 &  & SAIL~\cite{sail2025} & \underline{83.3} & \underline{39.0} & \underline{49.7} & 9.2 & 26.8 & \underline{41.6} \\
 &  & FreezeAlign~\cite{freezealign} & 78.7 & 28.3 & 32.0 & 3.9 & 29.7 & 34.5 \\
 &  & \cellcolor{relatehighlight}\textbf{\methodshort{} (ours)} & \cellcolor{relatehighlight}\textbf{96.7} & \cellcolor{relatehighlight}\textbf{47.2} & \cellcolor{relatehighlight}\textbf{58.3} & \cellcolor{relatehighlight}\textbf{14.0} & \cellcolor{relatehighlight}\textbf{34.5} & \cellcolor{relatehighlight}\textbf{50.1} \\
\cmidrule(lr){2-9}
 & \multirow{5}{*}{gte-Qwen2-1.5B} & CSA~\cite{csa2025} & 76.4 & 41.6 & 33.9 & 9.0 & 18.4 & 35.9 \\
 &  & STRUCTURE~\cite{structure2025} & \underline{93.5} & \underline{48.3} & 54.9 & \underline{19.5} & 32.9 & 49.8 \\
 &  & SAIL~\cite{sail2025} & 86.1 & 45.9 & \underline{56.5} & 15.8 & \underline{34.9} & 47.8 \\
 &  & FreezeAlign~\cite{freezealign} & 91.0 & 47.0 & 55.2 & \textbf{20.5} & \textbf{37.4} & \underline{50.2} \\
 &  & \cellcolor{relatehighlight}\textbf{\methodshort{} (ours)} & \cellcolor{relatehighlight}\textbf{96.6} & \cellcolor{relatehighlight}\textbf{52.5} & \cellcolor{relatehighlight}\textbf{63.2} & \cellcolor{relatehighlight}18.8 & \cellcolor{relatehighlight}33.7 & \cellcolor{relatehighlight}\textbf{52.9} \\
\cmidrule(lr){2-9}
 & \multirow{5}{*}{all-mpnet-base-v2} & CSA~\cite{csa2025} & 70.2 & 32.4 & 23.0 & 11.9 & 22.9 & 32.1 \\
 &  & STRUCTURE~\cite{structure2025} & \underline{91.4} & \underline{46.0} & 56.0 & \textbf{15.2} & 34.1 & \underline{48.5} \\
 &  & SAIL~\cite{sail2025} & 88.2 & 44.8 & \underline{56.9} & 10.7 & \textbf{37.3} & 47.6 \\
 &  & FreezeAlign~\cite{freezealign} & 89.9 & 42.1 & 54.5 & 13.0 & \underline{36.9} & 47.3 \\
 &  & \cellcolor{relatehighlight}\textbf{\methodshort{} (ours)} & \cellcolor{relatehighlight}\textbf{96.6} & \cellcolor{relatehighlight}\textbf{48.2} & \cellcolor{relatehighlight}\textbf{61.5} & \cellcolor{relatehighlight}\underline{14.4} & \cellcolor{relatehighlight}34.7 & \cellcolor{relatehighlight}\textbf{51.1} \\
\midrule
\multirow{10}{*}{DINOv2 ViT-B} & \multirow{5}{*}{RoBERTa-Large} & CSA~\cite{csa2025} & 82.6 & 37.2 & 26.7 & 10.1 & 15.6 & 34.4 \\
 &  & STRUCTURE~\cite{structure2025} & \underline{96.2} & 43.0 & \underline{57.5} & \underline{15.9} & 22.6 & 47.0 \\
 &  & SAIL~\cite{sail2025} & 95.0 & \underline{44.5} & 56.8 & 13.6 & 26.4 & \underline{47.3} \\
 &  & FreezeAlign~\cite{freezealign} & 92.4 & 41.9 & 52.9 & 13.8 & \underline{28.7} & 45.9 \\
 &  & \cellcolor{relatehighlight}\textbf{\methodshort{} (ours)} & \cellcolor{relatehighlight}\textbf{96.3} & \cellcolor{relatehighlight}\textbf{45.3} & \cellcolor{relatehighlight}\textbf{59.1} & \cellcolor{relatehighlight}\textbf{18.8} & \cellcolor{relatehighlight}\textbf{29.4} & \cellcolor{relatehighlight}\textbf{49.8} \\
\cmidrule(lr){2-9}
 & \multirow{5}{*}{all-mpnet-base-v2} & CSA~\cite{csa2025} & 79.8 & 33.5 & 32.8 & 10.6 & 9.1 & 33.2 \\
 &  & STRUCTURE~\cite{structure2025} & \underline{95.5} & 43.4 & 57.8 & \underline{16.2} & 20.7 & 46.7 \\
 &  & SAIL~\cite{sail2025} & 92.8 & 42.5 & 55.4 & 13.3 & 32.3 & 47.3 \\
 &  & FreezeAlign~\cite{freezealign} & 95.2 & \underline{45.6} & \underline{59.6} & \underline{16.2} & \underline{33.3} & \underline{50.0} \\
 &  & \cellcolor{relatehighlight}\textbf{\methodshort{} (ours)} & \cellcolor{relatehighlight}\textbf{95.7} & \cellcolor{relatehighlight}\textbf{47.5} & \cellcolor{relatehighlight}\textbf{60.3} & \cellcolor{relatehighlight}\textbf{16.5} & \cellcolor{relatehighlight}\textbf{40.0} & \cellcolor{relatehighlight}\textbf{52.0} \\
\bottomrule
\end{tabularx}
\end{table}

%% file: appendix/tables/a06_encoder_pairs_retrieval.tex
\begin{table}[!ht]
\caption{Cross-modal retrieval across encoder pairs (R@1, \%). Best results within each encoder pair and column are in \textbf{bold}; second-best results are \underline{underlined}.}
\label{tab:supp_encoder_retrieval}
\centering
\footnotesize
\renewcommand{\arraystretch}{1.20}
\setlength{\tabcolsep}{3pt}
\begin{tabularx}{\textwidth}{p{0.13\textwidth} p{0.14\textwidth} l *{5}{>{\centering\arraybackslash}X}}
\toprule
\textbf{Vision Model} & \textbf{Language Model} & \textbf{Method} & Flickr30k I2T & Flickr30k T2I & COCO I2T & COCO T2I & \textbf{Avg.} \\
\midrule
\multirow{20}{*}{DINOv2 ViT-L} & \multirow{5}{*}{RoBERTa-Large} & CSA~\cite{csa2025} & 43.4 & 32.6 & 25.1 & 19.2 & 30.1 \\
 &  & STRUCTURE~\cite{structure2025} & 61.3 & 47.6 & 40.4 & 29.4 & 44.7 \\
 &  & SAIL~\cite{sail2025} & 51.7 & 40.9 & 38.3 & 29.1 & 40.0 \\
 &  & FreezeAlign~\cite{freezealign} & \underline{62.9} & \underline{48.9} & \underline{43.5} & \underline{32.8} & \underline{47.0} \\
 &  & \cellcolor{relatehighlight}\textbf{\methodshort{} (ours)} & \cellcolor{relatehighlight}\textbf{76.5} & \cellcolor{relatehighlight}\textbf{61.9} & \cellcolor{relatehighlight}\textbf{56.3} & \cellcolor{relatehighlight}\textbf{42.6} & \cellcolor{relatehighlight}\textbf{59.3} \\
\cmidrule(lr){2-8}
 & \multirow{5}{*}{Qwen2.5-0.5B} & CSA~\cite{csa2025} & 15.8 & 7.9 & 13.9 & 8.1 & 11.4 \\
 &  & STRUCTURE~\cite{structure2025} & 32.2 & 23.1 & 16.2 & 15.5 & 21.8 \\
 &  & SAIL~\cite{sail2025} & \underline{42.0} & \underline{27.1} & \underline{27.6} & \underline{19.2} & \underline{29.0} \\
 &  & FreezeAlign~\cite{freezealign} & 24.3 & 19.2 & 19.9 & 15.0 & 19.6 \\
 &  & \cellcolor{relatehighlight}\textbf{\methodshort{} (ours)} & \cellcolor{relatehighlight}\textbf{71.0} & \cellcolor{relatehighlight}\textbf{53.2} & \cellcolor{relatehighlight}\textbf{51.0} & \cellcolor{relatehighlight}\textbf{37.7} & \cellcolor{relatehighlight}\textbf{53.2} \\
\cmidrule(lr){2-8}
 & \multirow{5}{*}{gte-Qwen2-1.5B} & CSA~\cite{csa2025} & 43.1 & 30.2 & 24.3 & 15.4 & 28.3 \\
 &  & STRUCTURE~\cite{structure2025} & 58.2 & 43.6 & 35.1 & 25.9 & 40.7 \\
 &  & SAIL~\cite{sail2025} & 55.2 & 39.1 & 38.7 & 27.7 & 40.2 \\
 &  & FreezeAlign~\cite{freezealign} & \underline{60.2} & \underline{45.7} & \underline{40.3} & \underline{30.0} & \underline{44.0} \\
 &  & \cellcolor{relatehighlight}\textbf{\methodshort{} (ours)} & \cellcolor{relatehighlight}\textbf{74.6} & \cellcolor{relatehighlight}\textbf{59.0} & \cellcolor{relatehighlight}\textbf{53.7} & \cellcolor{relatehighlight}\textbf{40.2} & \cellcolor{relatehighlight}\textbf{56.9} \\
\cmidrule(lr){2-8}
 & \multirow{5}{*}{all-mpnet-base-v2} & CSA~\cite{csa2025} & 39.4 & 30.1 & 22.9 & 17.7 & 27.5 \\
 &  & STRUCTURE~\cite{structure2025} & \underline{59.3} & \underline{45.0} & 38.0 & 28.2 & \underline{42.6} \\
 &  & SAIL~\cite{sail2025} & 50.2 & 37.3 & 36.6 & 27.4 & 37.9 \\
 &  & FreezeAlign~\cite{freezealign} & 55.1 & 42.8 & \underline{38.4} & \underline{29.1} & 41.4 \\
 &  & \cellcolor{relatehighlight}\textbf{\methodshort{} (ours)} & \cellcolor{relatehighlight}\textbf{73.4} & \cellcolor{relatehighlight}\textbf{58.3} & \cellcolor{relatehighlight}\textbf{54.2} & \cellcolor{relatehighlight}\textbf{40.6} & \cellcolor{relatehighlight}\textbf{56.6} \\
\midrule
\multirow{10}{*}{DINOv2 ViT-B} & \multirow{5}{*}{RoBERTa-Large} & CSA~\cite{csa2025} & 43.3 & 30.7 & 24.5 & 18.2 & 29.2 \\
 &  & STRUCTURE~\cite{structure2025} & \underline{58.0} & \underline{47.0} & \underline{39.2} & 28.3 & \underline{43.1} \\
 &  & SAIL~\cite{sail2025} & 49.8 & 40.3 & 38.4 & 28.5 & 39.2 \\
 &  & FreezeAlign~\cite{freezealign} & 55.4 & 41.5 & 38.0 & \underline{29.0} & 41.0 \\
 &  & \cellcolor{relatehighlight}\textbf{\methodshort{} (ours)} & \cellcolor{relatehighlight}\textbf{71.6} & \cellcolor{relatehighlight}\textbf{58.2} & \cellcolor{relatehighlight}\textbf{55.6} & \cellcolor{relatehighlight}\textbf{40.9} & \cellcolor{relatehighlight}\textbf{56.6} \\
\cmidrule(lr){2-8}
 & \multirow{5}{*}{all-mpnet-base-v2} & CSA~\cite{csa2025} & 35.0 & 26.8 & 21.8 & 16.2 & 25.0 \\
 &  & STRUCTURE~\cite{structure2025} & 57.6 & 43.4 & 36.8 & 27.2 & 41.2 \\
 &  & SAIL~\cite{sail2025} & 48.3 & 36.7 & 35.2 & 26.9 & 36.8 \\
 &  & FreezeAlign~\cite{freezealign} & \underline{61.3} & \underline{47.5} & \underline{44.2} & \underline{32.5} & \underline{46.4} \\
 &  & \cellcolor{relatehighlight}\textbf{\methodshort{} (ours)} & \cellcolor{relatehighlight}\textbf{70.9} & \cellcolor{relatehighlight}\textbf{55.6} & \cellcolor{relatehighlight}\textbf{52.7} & \cellcolor{relatehighlight}\textbf{39.2} & \cellcolor{relatehighlight}\textbf{54.6} \\
\bottomrule
\end{tabularx}
\end{table}

%% file: appendix/tables/a07_encoder_pairs_segmentation.tex
\begin{table}[!ht]
\caption{Zero-shot segmentation across encoder pairs (mIoU-fg, \%). Best results within each encoder pair and column are in \textbf{bold}; second-best results are \underline{underlined}.}
\label{tab:supp_encoder_segmentation}
\centering
\footnotesize
\renewcommand{\arraystretch}{1.20}
\setlength{\tabcolsep}{3pt}
\begin{tabularx}{\textwidth}{p{0.13\textwidth} p{0.14\textwidth} l *{4}{>{\centering\arraybackslash}X}}
\toprule
\textbf{Vision Model} & \textbf{Language Model} & \textbf{Method} & VOC2012 & Context & ADE20K & \textbf{Avg.} \\
\midrule
\multirow{20}{*}{DINOv2 ViT-L} & \multirow{5}{*}{RoBERTa-Large} & CSA~\cite{csa2025} & 6.3 & 3.2 & 2.3 & 3.9 \\
 &  & STRUCTURE~\cite{structure2025} & 11.8 & 8.2 & 3.8 & 7.9 \\
 &  & SAIL~\cite{sail2025} & \underline{22.3} & 21.0 & 10.7 & 18.0 \\
 &  & FreezeAlign~\cite{freezealign} & 20.2 & \underline{22.6} & \underline{12.0} & \underline{18.2} \\
 &  & \cellcolor{relatehighlight}\textbf{\methodshort{} (ours)} & \cellcolor{relatehighlight}\textbf{32.1} & \cellcolor{relatehighlight}\textbf{25.4} & \cellcolor{relatehighlight}\textbf{13.8} & \cellcolor{relatehighlight}\textbf{23.8} \\
\cmidrule(lr){2-7}
 & \multirow{5}{*}{Qwen2.5-0.5B} & CSA~\cite{csa2025} & 4.2 & 1.8 & 0.8 & 2.3 \\
 &  & STRUCTURE~\cite{structure2025} & 9.7 & 6.4 & 2.5 & 6.2 \\
 &  & SAIL~\cite{sail2025} & \underline{20.5} & \underline{16.1} & \underline{7.9} & \underline{14.8} \\
 &  & FreezeAlign~\cite{freezealign} & 14.3 & 11.0 & 3.7 & 9.7 \\
 &  & \cellcolor{relatehighlight}\textbf{\methodshort{} (ours)} & \cellcolor{relatehighlight}\textbf{38.8} & \cellcolor{relatehighlight}\textbf{17.3} & \cellcolor{relatehighlight}\textbf{10.5} & \cellcolor{relatehighlight}\textbf{22.2} \\
\cmidrule(lr){2-7}
 & \multirow{5}{*}{gte-Qwen2-1.5B} & CSA~\cite{csa2025} & 6.1 & 2.9 & 1.9 & 3.6 \\
 &  & STRUCTURE~\cite{structure2025} & 13.9 & 9.3 & 4.8 & 9.3 \\
 &  & SAIL~\cite{sail2025} & \underline{21.6} & 19.6 & 11.1 & \underline{17.4} \\
 &  & FreezeAlign~\cite{freezealign} & 18.4 & \underline{20.5} & \underline{11.7} & 16.9 \\
 &  & \cellcolor{relatehighlight}\textbf{\methodshort{} (ours)} & \cellcolor{relatehighlight}\textbf{36.4} & \cellcolor{relatehighlight}\textbf{28.0} & \cellcolor{relatehighlight}\textbf{16.5} & \cellcolor{relatehighlight}\textbf{27.0} \\
\cmidrule(lr){2-7}
 & \multirow{5}{*}{all-mpnet-base-v2} & CSA~\cite{csa2025} & 4.2 & 3.0 & 1.3 & 2.8 \\
 &  & STRUCTURE~\cite{structure2025} & 11.1 & 7.1 & 3.7 & 7.3 \\
 &  & SAIL~\cite{sail2025} & \underline{21.9} & 20.5 & \underline{10.8} & \underline{17.7} \\
 &  & FreezeAlign~\cite{freezealign} & 19.1 & \underline{20.7} & 10.6 & 16.8 \\
 &  & \cellcolor{relatehighlight}\textbf{\methodshort{} (ours)} & \cellcolor{relatehighlight}\textbf{35.7} & \cellcolor{relatehighlight}\textbf{22.4} & \cellcolor{relatehighlight}\textbf{13.4} & \cellcolor{relatehighlight}\textbf{23.8} \\
\midrule
\multirow{10}{*}{DINOv2 ViT-B} & \multirow{5}{*}{RoBERTa-Large} & CSA~\cite{csa2025} & 21.0 & 15.8 & 8.6 & 15.1 \\
 &  & STRUCTURE~\cite{structure2025} & \underline{23.4} & \underline{25.7} & \underline{12.0} & \underline{20.4} \\
 &  & SAIL~\cite{sail2025} & 21.9 & 21.6 & 11.4 & 18.3 \\
 &  & FreezeAlign~\cite{freezealign} & 21.3 & 22.9 & 11.5 & 18.6 \\
 &  & \cellcolor{relatehighlight}\textbf{\methodshort{} (ours)} & \cellcolor{relatehighlight}\textbf{29.0} & \cellcolor{relatehighlight}\textbf{28.4} & \cellcolor{relatehighlight}\textbf{15.2} & \cellcolor{relatehighlight}\textbf{24.2} \\
\cmidrule(lr){2-7}
 & \multirow{5}{*}{all-mpnet-base-v2} & CSA~\cite{csa2025} & 19.1 & 13.3 & 7.3 & 13.3 \\
 &  & STRUCTURE~\cite{structure2025} & \underline{23.3} & \textbf{25.2} & 12.2 & \underline{20.2} \\
 &  & SAIL~\cite{sail2025} & 21.8 & 21.3 & 11.4 & 18.2 \\
 &  & FreezeAlign~\cite{freezealign} & 21.9 & 24.4 & \underline{13.2} & 19.8 \\
 &  & \cellcolor{relatehighlight}\textbf{\methodshort{} (ours)} & \cellcolor{relatehighlight}\textbf{43.2} & \cellcolor{relatehighlight}\underline{24.9} & \cellcolor{relatehighlight}\textbf{13.8} & \cellcolor{relatehighlight}\textbf{27.3} \\
\bottomrule
\end{tabularx}
\end{table}

%% file: appendix/tables/a08_encoder_pairs_grounding.tex
\begin{table}[!ht]
\caption{Phrase grounding across encoder pairs (pointing accuracy, \%). Best results within each encoder pair and column are in \textbf{bold}; second-best results are \underline{underlined}.}
\label{tab:supp_encoder_grounding}
\centering
\footnotesize
\renewcommand{\arraystretch}{1.20}
\setlength{\tabcolsep}{3pt}
\begin{tabularx}{\textwidth}{p{0.13\textwidth} p{0.14\textwidth} l *{4}{>{\centering\arraybackslash}X}}
\toprule
\textbf{Vision Model} & \textbf{Language Model} & \textbf{Method} & Flickr30k & ReferItGame & RefCOCOg & \textbf{Avg.} \\
\midrule
\multirow{20}{*}{DINOv2 ViT-L} & \multirow{5}{*}{RoBERTa-Large} & CSA~\cite{csa2025} & 35.3 & 26.0 & 24.9 & 28.7 \\
 &  & STRUCTURE~\cite{structure2025} & 38.7 & 31.2 & 34.1 & 34.7 \\
 &  & SAIL~\cite{sail2025} & 55.8 & 44.3 & 44.2 & 48.1 \\
 &  & FreezeAlign~\cite{freezealign} & \underline{60.6} & \underline{48.2} & \underline{48.7} & \underline{52.5} \\
 &  & \cellcolor{relatehighlight}\textbf{\methodshort{} (ours)} & \cellcolor{relatehighlight}\textbf{69.1} & \cellcolor{relatehighlight}\textbf{50.5} & \cellcolor{relatehighlight}\textbf{56.6} & \cellcolor{relatehighlight}\textbf{58.7} \\
\cmidrule(lr){2-7}
 & \multirow{5}{*}{Qwen2.5-0.5B} & CSA~\cite{csa2025} & 29.8 & 25.9 & 24.9 & 26.9 \\
 &  & STRUCTURE~\cite{structure2025} & 33.8 & 27.3 & 30.6 & 30.6 \\
 &  & SAIL~\cite{sail2025} & 44.7 & \underline{33.9} & 41.5 & 40.0 \\
 &  & FreezeAlign~\cite{freezealign} & \underline{52.0} & 32.4 & \underline{42.0} & \underline{42.1} \\
 &  & \cellcolor{relatehighlight}\textbf{\methodshort{} (ours)} & \cellcolor{relatehighlight}\textbf{53.4} & \cellcolor{relatehighlight}\textbf{39.6} & \cellcolor{relatehighlight}\textbf{54.2} & \cellcolor{relatehighlight}\textbf{49.1} \\
\cmidrule(lr){2-7}
 & \multirow{5}{*}{gte-Qwen2-1.5B} & CSA~\cite{csa2025} & 27.2 & 23.8 & 17.8 & 22.9 \\
 &  & STRUCTURE~\cite{structure2025} & 35.2 & 26.3 & 31.5 & 31.0 \\
 &  & SAIL~\cite{sail2025} & 47.7 & 34.6 & 41.7 & 41.3 \\
 &  & FreezeAlign~\cite{freezealign} & \underline{55.8} & \underline{40.5} & \underline{48.0} & \underline{48.1} \\
 &  & \cellcolor{relatehighlight}\textbf{\methodshort{} (ours)} & \cellcolor{relatehighlight}\textbf{57.1} & \cellcolor{relatehighlight}\textbf{40.7} & \cellcolor{relatehighlight}\textbf{55.1} & \cellcolor{relatehighlight}\textbf{50.9} \\
\cmidrule(lr){2-7}
 & \multirow{5}{*}{all-mpnet-base-v2} & CSA~\cite{csa2025} & 43.9 & 28.0 & 34.2 & 35.4 \\
 &  & STRUCTURE~\cite{structure2025} & 40.7 & 29.8 & 34.7 & 35.1 \\
 &  & SAIL~\cite{sail2025} & 55.0 & 43.0 & 44.0 & 47.3 \\
 &  & FreezeAlign~\cite{freezealign} & \underline{59.5} & \underline{46.3} & \underline{46.8} & \underline{50.9} \\
 &  & \cellcolor{relatehighlight}\textbf{\methodshort{} (ours)} & \cellcolor{relatehighlight}\textbf{67.7} & \cellcolor{relatehighlight}\textbf{49.3} & \cellcolor{relatehighlight}\textbf{56.3} & \cellcolor{relatehighlight}\textbf{57.8} \\
\midrule
\multirow{10}{*}{DINOv2 ViT-B} & \multirow{5}{*}{RoBERTa-Large} & CSA~\cite{csa2025} & 21.0 & 24.2 & 14.3 & 19.8 \\
 &  & STRUCTURE~\cite{structure2025} & 58.4 & 48.4 & \underline{50.7} & \underline{52.5} \\
 &  & SAIL~\cite{sail2025} & \underline{58.6} & 48.4 & 45.3 & 50.7 \\
 &  & FreezeAlign~\cite{freezealign} & 57.8 & \underline{49.7} & 46.7 & 51.4 \\
 &  & \cellcolor{relatehighlight}\textbf{\methodshort{} (ours)} & \cellcolor{relatehighlight}\textbf{70.4} & \cellcolor{relatehighlight}\textbf{53.1} & \cellcolor{relatehighlight}\textbf{59.3} & \cellcolor{relatehighlight}\textbf{60.9} \\
\cmidrule(lr){2-7}
 & \multirow{5}{*}{all-mpnet-base-v2} & CSA~\cite{csa2025} & 40.7 & 27.8 & 32.4 & 33.6 \\
 &  & STRUCTURE~\cite{structure2025} & 59.1 & 48.8 & 51.2 & 53.0 \\
 &  & SAIL~\cite{sail2025} & 57.4 & 48.8 & 45.2 & 50.5 \\
 &  & FreezeAlign~\cite{freezealign} & \underline{60.7} & \underline{49.3} & \underline{51.5} & \underline{53.9} \\
 &  & \cellcolor{relatehighlight}\textbf{\methodshort{} (ours)} & \cellcolor{relatehighlight}\textbf{69.6} & \cellcolor{relatehighlight}\textbf{51.7} & \cellcolor{relatehighlight}\textbf{57.5} & \cellcolor{relatehighlight}\textbf{59.6} \\
\bottomrule
\end{tabularx}
\end{table}

%% file: appendix/sections/e_pathology_classification.tex
\begin{appendixtext}
\section{Additional Pathology Classification Results}
\label{sec:supp_pathology_prompts}

\vspace{8pt}
\subsection{Data, Encoders, and Optimization}
\label{app:pathology_training}
Post-hoc methods use frozen UNI~\cite{chen2024uni} (block 23; $224\times224$ images giving 196 spatial patches and a CLS token) and PubMedBERT~\cite{gu2021pubmedbert} (hidden layer 12) encoders. PathCap~\cite{sun2023pathasst} pairs figure-panel images with captions. Of 223,169 records, 1,351 captions longer than 128 tokens are removed, and the remaining 221,818 pairs are split with seed 42 into 220,318 training and 1,500 validation pairs with no shared images. Training follows Section~\ref{sec:supp_experimental_setup} with pooling temperature $\tau_p=0.05$ and contrastive temperature $\tau=0.05$. STRUCTURE, SAIL, FreezeAlign, and \methodshort{} report mean and standard deviation over three seeds; CSA and the VLM references use single evaluations.

\subsection{Evaluation and Prompting}
\label{app:pathology_evaluation}
CRC100K~\cite{kather2019crc} is evaluated on CRC-VAL-HE-7K (7,180 images), SICAP~\cite{silva2020sicap} on the SICAPv2 test partition (2,122 images), and LC25000~\cite{borkowski2019lc25000} on all 25,000 images; LC25000-Lung (15,000 images) and LC25000-Colon (10,000 images) are its organ-specific subsets. Balanced accuracy averages recall over classes, weighted F1 averages class-wise F1 scores by class support, and SICAP additionally reports quadratic-weighted Cohen's $\kappa$ over the ordinal Gleason scale. All datasets use the 22 histopathology templates listed below: each prompt representation is normalized, averaged over templates and available synonyms, and renormalized. CRC100K uses the class-name synonyms released with CONCH~\cite{lu2024conch}; the other datasets use one class name per class.

The class names for CRC100K are adipose, background, debris, lymphocytes, mucus, smooth muscle, normal colon mucosa, cancer-associated stroma, and colorectal adenocarcinoma epithelium. SICAP uses benign prostate glands and prostatic adenocarcinoma with Gleason patterns 3, 4, and 5. LC25000-Colon uses colorectal adenocarcinoma and normal colon mucosa; LC25000-Lung uses lung adenocarcinoma, benign lung tissue, and lung squamous cell carcinoma. The full LC25000 task combines these five names. SICAP class names and the benign-lung wording are author-defined; the CRC100K synonyms follow the released CONCH set.

\vspace{8pt}
\begin{itemize}\setlength{\itemsep}{4pt}
\item \textbf{ADI}: adipose; adipose tissue; adipocytes; fat; fat cells.
\item \textbf{BACK}: background; penmarking; empty space; background artifacts.
\item \textbf{DEB}: debris; colorectal adenocarcinoma debris and necrosis; necrosis; necrotic debris.
\item \textbf{LYM}: lymphocytes; lymphoid aggregate; immune cells; lymphoid infiltrate; inflammatory cells.
\item \textbf{MUC}: mucus; mucin; mucus pool; mucin pool.
\item \textbf{MUS}: smooth muscle; smooth muscle tissue; muscle; muscularis propria; muscularis mucosa.
\item \textbf{NORM}: normal colon mucosa; uninvolved colon mucosa; normal colonic mucosa; benign epithelium.
\item \textbf{STR}: cancer-associated stroma; tumor-associated stroma; stromal cells; stromal tissue; stroma.
\item \textbf{TUM}: colorectal adenocarcinoma epithelium; colorectal adenocarcinoma; tumor; adenocarcinoma; malignant epithelium.
\end{itemize}
\vspace{12pt}

All pathology benchmarks use the following 22 templates from the CONCH prompt set~\cite{lu2024conch}.

\vspace{8pt}
\begin{itemize}\setlength{\itemsep}{4pt}
\item \textbf{Histopathology} (22 templates).\par
{\footnotesize\raggedright \texttt{\{class\}.}
\quad \texttt{a photomicrograph showing \{class\}.}
\quad \texttt{a photomicrograph of \{class\}.}
\quad \texttt{an image of \{class\}.}
\quad \texttt{an image showing \{class\}.}
\quad \texttt{an example of \{class\}.}
\quad \texttt{\{class\} is shown.}
\quad \texttt{this is \{class\}.}
\quad \texttt{there is \{class\}.}
\quad \texttt{a histopathological image showing \{class\}.}
\quad \texttt{a histopathological image of \{class\}.}
\quad \texttt{a histopathological photograph of \{class\}.}
\quad \texttt{a histopathological photograph showing \{class\}.}
\quad \texttt{an H\&E stained image of \{class\}.}
\quad \texttt{an H\&E stained image showing \{class\}.}
\quad \texttt{an H\&E image showing \{class\}.}
\quad \texttt{an H\&E image of \{class\}.}
\quad \texttt{\{class\}, H\&E stain.}
\quad \texttt{\{class\}, H\&E.}
\quad \texttt{shows \{class\}.}
\quad \texttt{presence of \{class\}.}
\quad \texttt{\{class\} is present.}\par}
\end{itemize}

\subsection{VLM Reference Configurations}
\label{app:pathology_vlms}
The VLM references use their released weights without task-specific updates and the same class descriptions and template protocol as above.

\vspace{8pt}
\begin{itemize}\setlength{\itemsep}{4pt}
\item \textbf{CLIP} (ICML 2021)~\cite{radford2021learning} is a general-domain VLM trained contrastively on 400M web image--text pairs. We use OpenCLIP ViT-B/16 with OpenAI weights at $224\times224$.
\item \textbf{QuiltNet} (NeurIPS 2023)~\cite{ikezogwo2023quilt} fine-tunes CLIP on Quilt-1M, a histopathology image--text dataset curated from educational videos. We use QuiltNet-B/32 at $224\times224$.
\item \textbf{PLIP} (Nature Medicine 2023)~\cite{huang2023plip} fine-tunes CLIP on pathology image--text pairs collected from medical Twitter. We use the released checkpoint at $224\times224$ with a 77-token text limit.
\item \textbf{CONCH} (Nature Medicine 2024)~\cite{lu2024conch} is a pathology vision--language foundation model pretrained on 1.17M histopathology image--caption pairs. We use the ViT-B/16 checkpoint at $224\times224$ for CRC100K and SICAP and at $448\times448$ for the LC25000 tasks.
\item \textbf{MUSK} (Nature 2025)~\cite{xiang2025musk} is a pathology vision--language foundation model pretrained on large-scale unpaired pathology images and text and then aligned with image--text pairs. We use the large patch-16 checkpoint at $384\times384$ with bicubic resizing, center cropping, and Inception-style normalization; text uses its XLM-RoBERTa tokenizer with a 100-token limit.
\end{itemize}

\subsection{Scaling with Paired Pathology Data}
\label{app:pathology_scaling}
Fig.~\ref{fig:supp_pathology_scaling} reports the scaling of \methodshort{} on the five pathology benchmarks at 1K, 5K, 20K, 100K, and the full 220,318 PathCap pairs (seed 42). Subsets are sampled randomly, all sizes share the 1,500-pair validation set, and the learning rate is selected at each size. The prompt protocol is unchanged. Increasing the paired alignment set improves performance overall on all five pathology benchmarks.
\end{appendixtext}

\begin{figure}[!ht]
\centering
\begin{minipage}{0.8\textwidth}
\centering
\includegraphics[width=0.6125\linewidth]{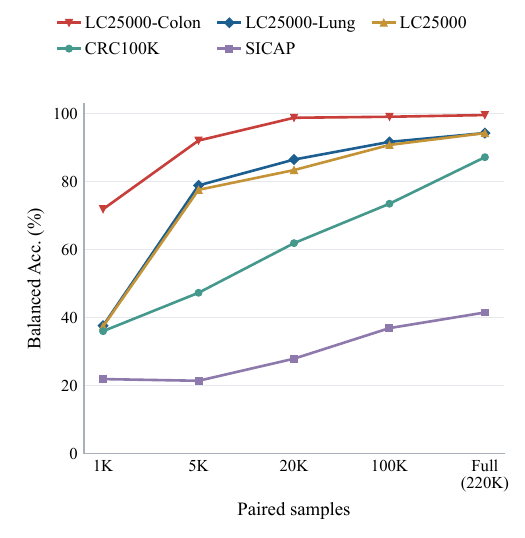}
\caption{Zero-shot pathology classification as the paired alignment set increases from 1K to 220K pairs. Curves show balanced accuracy (\%) using seed 42. LC25000-Lung and LC25000-Colon are subsets of LC25000. All datasets use 22 histopathology templates; CRC100K additionally uses CONCH class-name synonyms.}
\label{fig:supp_pathology_scaling}
\end{minipage}
\end{figure}

\clearpage
\begin{appendixtext}
\subsection{Additional Classification Results}
\label{app:pathology_results}
Table~\ref{tab:supp_pathology_prompts} reports balanced accuracy and weighted F1 for CRC100K and SICAP, together with quadratic-weighted Cohen's $\kappa$ for SICAP. Table~\ref{tab:supp_pathology_lc25000} reports balanced accuracy and weighted F1 for LC25000 and its lung and colon subsets.
\end{appendixtext}

\input{appendix/tables/a09_pathology_crc100k_sicap}
\input{appendix/tables/a10_pathology_lc25000}

%% file: appendix/tables/a09_pathology_crc100k_sicap.tex
\begin{table}[!ht]
\centering
\begin{minipage}{0.8\textwidth}
\caption{Zero-shot classification on CRC100K and SICAP. Balanced accuracy and weighted F1 are percentages; SICAP's quadratic-weighted Cohen's $\kappa$ is reported on its original scale. Trainable post-hoc methods report mean $\pm$ standard deviation over three seeds; CSA and VLMs use single evaluations. All datasets use 22 histopathology templates; CRC100K additionally uses CONCH class-name synonyms. Best and second-best results across all methods are in \textbf{bold} and \underline{underlined}, respectively.}
\label{tab:supp_pathology_prompts}
\centering
\footnotesize
\renewcommand{\arraystretch}{1.20}
\setlength{\tabcolsep}{2pt}
\newlength{\pathologyscorewidthNine}
\settowidth{\pathologyscorewidthNine}{\textbf{99.9}\,$\pm$\,9.9}
\newlength{\pathologykappawidthNine}
\settowidth{\pathologykappawidthNine}{\textbf{0.999}\,$\pm$\,0.999}
\begin{tabularx}{\linewidth}{l *{4}{>{\raggedright\arraybackslash\hspace*{\dimexpr(\hsize-\pathologyscorewidthNine)/2\relax}}X} >{\raggedright\arraybackslash\hspace*{\dimexpr(\hsize-\pathologykappawidthNine)/2\relax}}X}
\toprule
& \multicolumn{2}{c}{\textbf{CRC100K}} & \multicolumn{3}{c}{\textbf{SICAP}} \\
\cmidrule(lr){2-3}\cmidrule(lr){4-6}
\textbf{Method} & \multicolumn{1}{c}{Balanced Acc.} & \multicolumn{1}{c}{Weighted F1} & \multicolumn{1}{c}{Balanced Acc.} & \multicolumn{1}{c}{Weighted F1} & \multicolumn{1}{c}{Weighted $\kappa$} \\
\midrule
CLIP~\cite{radford2021learning} & 27.1 & 24.7 & 20.2 & 19.2 & -0.026 \\
\midrule
QuiltNet~\cite{ikezogwo2023quilt} & 59.8 & 59.9 & 30.9 & 19.8 & 0.090 \\
PLIP~\cite{huang2023plip} & 67.4 & 68.7 & 27.1 & 27.1 & 0.176 \\
CONCH~\cite{lu2024conch} & 79.1 & 80.3 & 35.0 & 25.5 & 0.352 \\
MUSK~\cite{xiang2025musk} & \underline{79.9} & \underline{83.7} & \underline{37.2} & 27.6 & \textbf{0.423} \\
\midrule
CSA~\cite{csa2025} & 41.4 & 40.1 & 25.5 & 11.2 & -0.030 \\
STRUCTURE~\cite{structure2025} & 59.2\,$\pm$\,1.1 & 54.2\,$\pm$\,1.0 & 34.8\,$\pm$\,1.1 & \textbf{38.7}\,$\pm$\,1.2 & 0.394\,$\pm$\,0.047 \\
SAIL~\cite{sail2025} & 63.5\,$\pm$\,2.6 & 61.1\,$\pm$\,3.6 & \underline{37.2}\,$\pm$\,4.0 & \underline{35.7}\,$\pm$\,4.1 & 0.315\,$\pm$\,0.148 \\
FreezeAlign~\cite{freezealign} & 74.8\,$\pm$\,6.2 & 71.7\,$\pm$\,8.5 & 36.2\,$\pm$\,3.2 & 29.3\,$\pm$\,5.2 & 0.357\,$\pm$\,0.068 \\
\midrule
\rowcolor{relatehighlight}
\textbf{\methodshort{} (ours)} & \textbf{85.6}\,$\pm$\,1.9 & \textbf{88.2}\,$\pm$\,1.6 & \textbf{38.3}\,$\pm$\,3.3 & 31.6\,$\pm$\,3.6 & \underline{0.405}\,$\pm$\,0.018 \\
\bottomrule
\end{tabularx}
\end{minipage}
\end{table}

%% file: appendix/tables/a10_pathology_lc25000.tex
\begin{table}[!ht]
\centering
\begin{minipage}{0.8\textwidth}
\caption{Zero-shot classification on LC25000 and its lung and colon subsets (\%). Trainable post-hoc methods report mean $\pm$ standard deviation over three seeds; CSA and VLMs use single evaluations. All datasets use 22 histopathology templates; CRC100K additionally uses CONCH class-name synonyms. Best and second-best results across all methods are in \textbf{bold} and \underline{underlined}, respectively.}
\label{tab:supp_pathology_lc25000}
\centering
\footnotesize
\renewcommand{\arraystretch}{1.20}
\setlength{\tabcolsep}{2pt}
\newlength{\pathologyscorewidthTen}
\settowidth{\pathologyscorewidthTen}{\textbf{99.9}\,$\pm$\,9.9}
\begin{tabularx}{\linewidth}{l *{6}{>{\raggedright\arraybackslash\hspace*{\dimexpr(\hsize-\pathologyscorewidthTen)/2\relax}}X}}
\toprule
& \multicolumn{2}{c}{\textbf{LC25000}} & \multicolumn{2}{c}{\textbf{LC25000-Lung}} & \multicolumn{2}{c}{\textbf{LC25000-Colon}} \\
\cmidrule(lr){2-3}\cmidrule(lr){4-5}\cmidrule(lr){6-7}
\textbf{Method} & \multicolumn{1}{c}{Balanced Acc.} & \multicolumn{1}{c}{Weighted F1} & \multicolumn{1}{c}{Balanced Acc.} & \multicolumn{1}{c}{Weighted F1} & \multicolumn{1}{c}{Balanced Acc.} & \multicolumn{1}{c}{Weighted F1} \\
\midrule
CLIP~\cite{radford2021learning} & 27.7 & 24.3 & 41.6 & 40.2 & 53.3 & 40.4 \\
\midrule
QuiltNet~\cite{ikezogwo2023quilt} & 73.2 & 72.7 & 78.1 & 77.3 & 94.1 & 94.1 \\
PLIP~\cite{huang2023plip} & 77.4 & 78.1 & 85.8 & 86.0 & 90.5 & 90.4 \\
CONCH~\cite{lu2024conch} & \underline{92.9} & \underline{92.8} & \textbf{94.8} & \textbf{94.8} & \underline{99.6} & \underline{99.6} \\
MUSK~\cite{xiang2025musk} & 88.8 & 88.6 & 91.8 & 91.6 & \textbf{99.7} & \textbf{99.7} \\
\midrule
CSA~\cite{csa2025} & 56.9 & 53.8 & 46.8 & 42.7 & 88.8 & 88.7 \\
STRUCTURE~\cite{structure2025} & 64.0\,$\pm$\,1.3 & 61.4\,$\pm$\,1.8 & 66.0\,$\pm$\,3.6 & 63.8\,$\pm$\,4.4 & 93.0\,$\pm$\,2.9 & 93.0\,$\pm$\,2.9 \\
SAIL~\cite{sail2025} & 81.2\,$\pm$\,3.4 & 79.4\,$\pm$\,2.7 & 76.6\,$\pm$\,5.4 & 74.9\,$\pm$\,4.6 & 98.9\,$\pm$\,0.8 & 98.9\,$\pm$\,0.8 \\
FreezeAlign~\cite{freezealign} & 86.0\,$\pm$\,2.0 & 85.9\,$\pm$\,1.8 & 82.0\,$\pm$\,3.1 & 82.0\,$\pm$\,2.8 & 99.1\,$\pm$\,0.4 & 99.1\,$\pm$\,0.4 \\
\midrule
\rowcolor{relatehighlight}
\textbf{\methodshort{} (ours)} & \textbf{93.6}\,$\pm$\,1.1 & \textbf{93.4}\,$\pm$\,1.3 & \underline{93.8}\,$\pm$\,0.8 & \underline{93.8}\,$\pm$\,0.8 & \underline{99.6}\,$\pm$\,0.2 & \underline{99.6}\,$\pm$\,0.2 \\
\bottomrule
\end{tabularx}
\end{minipage}
\end{table}

%% file: appendix/sections/f_cross_modal_anchor_consistency.tex
\begin{figure}[!ht]
\centering
\captionsetup[subfloat]{font={normalsize,rm},labelfont=rm,textfont=rm}
\subfloat[Flickr30k]{\includegraphics[width=0.49\textwidth]{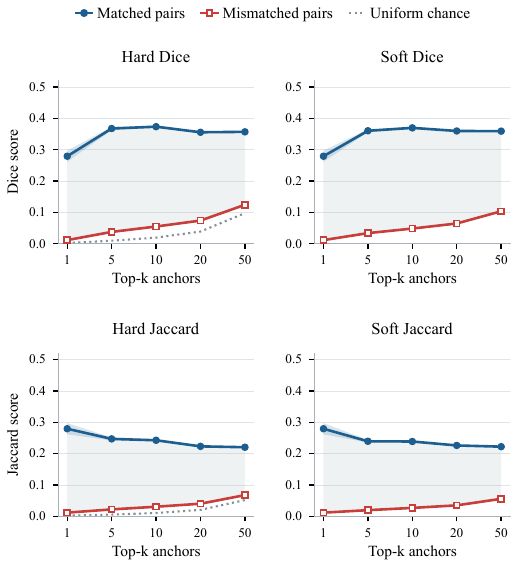}}\hfill
\subfloat[COCO]{\includegraphics[width=0.49\textwidth]{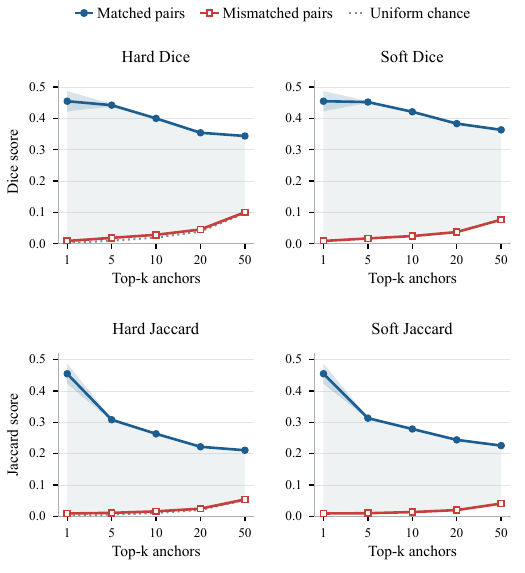}}
\caption{Cross-modal anchor consistency across top-$k$ selections ($K{=}512$). Each dataset subfigure contains Hard Dice and Soft Dice in the first row and Hard Jaccard and Soft Jaccard in the second row. Lines and colored bands show mean and standard deviation over three seeds; light gray shading highlights the matched--mismatched gap. Dotted gray lines show the expected overlap of independently and uniformly selected anchor sets for the hard metrics.}
\label{fig:supp_anchor_overlap}
\end{figure}

\begin{appendixtext}
\section{Cross-Modal Anchor Consistency}
\label{app:anchor_overlap}

\vspace{8pt}
We evaluate the $K{=}512$ token-level ASP models trained with seeds 42, 44, and 123 on Flickr30k test and COCO Karpathy test. For each image--caption pair, we independently select the top-$k$ coordinates of the pooled representations $\mathbf h^v$ and $\mathbf h^t$. Evaluation is caption-level: the same image may occur with multiple captions. Mismatched scores use five randomly sampled non-matching captions per image, with sampling seed 42.

For top-$k$ anchor sets $A$ and $B$, hard Dice is $2|A\cap B|/(|A|+|B|)=|A\cap B|/k$, and hard Jaccard is $|A\cap B|/|A\cup B|$. For soft metrics, each profile is restricted to its own top-$k$ support, negative entries are clamped to zero, and the truncated vector is L1-normalized. Denoting the resulting nonnegative vectors by $u$ and $v$, soft Dice is $2\sum_j\min(u_j,v_j)/(\sum_j u_j+\sum_j v_j)$, and soft Jaccard is $\sum_j\min(u_j,v_j)/\sum_j\max(u_j,v_j)$.

\vspace{4pt}
Fig.~\ref{fig:supp_anchor_overlap} shows all four metrics for $k\in\{1,5,10,20,50\}$. Dice and Jaccard satisfy $J=D/(2-D)$ for each pair. For independently and uniformly selected size-$k$ anchor sets, the intersection size $I$ follows a hypergeometric distribution. The hard Dice chance is $k/512$, and the hard Jaccard chance is $\mathbb{E}[I/(2k-I)]$. No uniform chance curve is shown for the soft metrics, whose expectations also depend on the anchor-weight distributions.
\end{appendixtext}

